%% file: paper.tex
\documentclass[preprint,review,12pt,times,3p]{elsarticle}

\input{_packages}

\usepackage[some]{background}
\usepackage{lipsum}

\journal{Journal of Parallel and Distributed Computing}

\SetBgContents{\textcolor{black}{Preprint. https://doi.org/10.1016/j.jpdc.2025.105053}}
\SetBgScale{1}
\SetBgAngle{0}
\SetBgOpacity{1} 
\SetBgPosition{current page.north east}
\SetBgHshift{-7cm}
\SetBgVshift{-1cm}

\begin{document}
\BgThispage
\begin{frontmatter}

\title{GPU Memory Usage Optimization for Backward Propagation in Deep Network Training}

%\cortext[cor1]{Corresponding author. E-mail: dyhong@iis.sinica.edu.tw. Telephone number: +886-2-27883799 ext. 1818.}

\author[inst2]{Ding-Yong Hong}
\author[inst1]{Tzu-Hsien Tsai}
\author[inst1]{Ning Wang}
\author[inst1]{Pangfeng Liu}
\author[inst2]{Jan-Jan Wu}

\affiliation[inst2]{organization={Institute of Information Science, Academia Sinica},%Department and Organization
            city={Taipei},
            country={Taiwan}}
\affiliation[inst1]{organization={Department of Computer Science and Information Engineering, National Taiwan University},%Department and Organization
            city={Taipei},
            country={Taiwan}}

\input{0_abstract}


\end{frontmatter}

\input{1_introduction}
\input{2_related}
\input{3_algorithm}
\input{4_dynamic_checkpoint}
\input{5_experiment}
\input{6_conclusion}

\bibliographystyle{elsarticle-harv} 
\bibliography{reference}

\end{document}

%% file: _packages.tex
\usepackage{amsthm,amssymb,amsmath}
\usepackage{balance}
\usepackage{enumerate}
\usepackage{graphicx}
\usepackage{algorithm}
\usepackage{algpseudocode}
\usepackage{lipsum,adjustbox}
\usepackage{pgfplots}\pgfplotsset{compat=1.17}
\usepackage[strings]{underscore}
\usepackage{tikz}
\usetikzlibrary{
  arrows.meta,
  calc,
  fit,
  positioning,
  patterns,
  quotes,
  colorbrewer
}

\usepackage{subcaption}
\newcommand{\LineLabel}[1]{\addtocounter{ALG@line}{-1}\refstepcounter{ALG@line}\label{#1}}  % Associates the line number with a label

\usepackage{bibentry}
\usepackage[numbers]{natbib}

\newcommand{\comment}[1]{}
\newtheorem{theorem}{Theorem}
\newtheorem{lemma}{Lemma}

\usepackage{amsmath}

\DeclareMathOperator*{\argmin}{argmin }

\usepackage{bbding} % for checkmarks

%% file: 0_abstract.tex
\begin{abstract}

In modern Deep Learning, it has been a trend to design larger Deep Neural Networks (DNNs) for the execution of more complex tasks and better accuracy. On the other hand, Convolutional Neural Networks (CNNs) have become the standard method for most of computer vision tasks.
However, the memory allocation for the intermediate data in convolution layers can cause severe memory pressure during model training.
Many solutions have been proposed to resolve the problem.
Besides hardware-dependent solutions, a general methodology {\em rematerialization} can reduce GPU memory usage by trading computation for memory efficiently.
The idea is to select a set of intermediate results during the forward phase as {\em checkpoints}, and only save them in memory to reduce memory usage.
The backward phase recomputes the intermediate data from the closest checkpoints in memory as needed.
This recomputation increases execution time but saves memory by not storing all intermediate results in memory during the forward phase.
In this paper, we will focus on efficiently finding the optimal checkpoint subset to achieve the least peak memory usage during the model training. We first describe the theoretical background of the training of a neural network using mathematical equations.
We use these equations to identify all essential data required during both forward and backward phases to compute the gradient of weights of the model.
We first identify the {\em checkpoint selection} problem and propose a dynamic programming algorithm with time complexity $O(n^3)$ to solve the problem of finding the optimal checkpoint subset.
With extensive experiments, we formulate a more accurate description of the problem using our theoretical analysis and revise the objective function based on the tracing, and propose an $O(n)$-time algorithm for finding the optimal checkpoint subset.

\end{abstract}

%% file: 1_introduction.tex
\section{Introduction}\label{sec:introduction}

\subsection{The Memory Pressure Problem in Deep Learning}

% The trend of model size in Deep Learning.

In Deep Learning, it has been a trend to design larger Deep Neural Networks (DNNs) for the execution of more complex tasks and better accuracy in the past few years~\cite{He2015DeepRL, 10.5555/3298023.3298188, Wu2016GooglesNM, 2016arXiv160507146Z, Zoph2017LearningTA, Simonyan2014VeryDC}. For example, the idea of the residual block behind ResNets~\cite{He2015DeepRL} makes it possible to train networks with over one thousand layers.
Among all kinds of models, Convolutional Neural Networks (CNN) have become the standard method for object detection and most other computer vision applications.
However, the memory allocated for the intermediate data of these convolution layers and the growth of the number of model parameters has become a problem~\cite{Wu2016HighperformanceSS, 8237506, Yu2022CoCaCC, Tan2019EfficientNetRM, Hu2017SqueezeandExcitationN}.
On the other hand, some research shows that a sufficiently large batch size is required for good and fast convergence~\cite{Pudipeddi2020TrainingLN}.
What makes it worse, there is a piece of evidence that the number of parameters in the state-of-the-art neural network has doubled around every 2.4 years~\cite{Goodfellow-et-al-2016}.

% Existing solutions and their limitations.

Many solutions have been proposed to solve the memory pressure problem~\cite{Huang2020SwapAdvisorPD, Rhu2016vDNNVD, Le2018TFLMSLM, Han2015DeepCC, Han2015LearningBW, Gupta2015DeepLW, 10.1145/2925426.2926294}.
%% with hardware requirement.
Offloading moves the activations during the forward phase from the memory of GPUs to the CPU memory,  and then prefetches the data back at the appropriate stage during the backward phase~\cite{Rhu2016vDNNVD}.
While manufacturers have designed high-end hardware, e.g. NVLink~\cite{Le2018TFLMSLM}, to support efficient communication between GPU and CPU, this approach may be too expensive for some researchers since NVLink is expensive.

Besides hardware-dependent solutions, there are approaches to reducing GPU memory usage by pruning unimportant connections~\cite{Han2015DeepCC, Han2015LearningBW}. We can also reduce the size of data by reducing the precision of floating point numbers. Some research shows that we can train DNN models using 16-bit wide fixed-point number with little or no degradation in classification accuracy~\cite{Gupta2015DeepLW}. More importantly, it enables larger models to fit within the given memory budget~\cite{10.1145/2925426.2926294}.

%% related methods: Trading computation for memory

In addition to hardware-dependent solutions, we have to use some techniques to keep up with the pace of the growing memory requirement.
The general methodology {\em trading computation for memory}~\cite{chen2016training} or {\em rematerialization}~\cite{Gruslys2016MemoryEfficientBT, kirisame2021dynamic, herrmann2019optimal} aims to save GPU memory by not storing the activations of certain layers during the forward phase
%delays the computation of activations of a subset of layers during the forward phase to save GPU memory
and recomputing them in batch during the backward phase.
The method eliminates the memory and hardware requirement for training large-scale DNN models without loss of accuracy at the cost of a moderate increase in time on recomputing activation data.
Since the idea of rematerialization and offloading are independent, another approach is to combine the two to reduce memory usage~\cite{Beaumont2021EfficientCO}.
On the other hand, for image recognition tasks, an improvement of the ResNet~\cite{He2015DeepRL} model has been proposed to reduce memory consumption by making the activations of most layers reconstructible from the activation of the next layer. This eliminates the need to store intermediate data during the forward phase~\cite{Gomez2017TheRR}.

Our work will focus on efficiently finding the optimal checkpoint subset to achieve the least peak memory usage during the model training. To the best of our knowledge, most existing works are built upon the {\em sublinear memory cost} method proposed by Chen et al.~\cite{chen2016training} since it is effective and easy to implement.
%But with our extensive experiments, there is a huge difference in the peak memory usage with and without the optimal checkpoint subset.
But with our extensive experiments, there is a huge difference in the peak memory usage with the checkpoint subset found by their method and by our method.
On the other hand, the {\em arbitrary computation graph} solver proposed by Feng et al.~\cite{feng2021optimal}  is designed to solve for the optimal checkpoint subset using the same objective function as the sublinear memory cost method. We will show that the objective function is not precise in describing the behavior of the state-of-the-art deep learning platform PyTorch and thus the algorithm does not give the optimal checkpoint subset. We will demonstrate the difference in the experiment section.

In this paper, we first describe the theoretical background of training a neural network using mathematical equations. We use these equations to identify all essential data required during both the forward and backward phases to compute the gradient of weights of the model. We summarize our analysis with a piece of pseudocode to represent the implementation of existing deep learning platforms for ease of identifying the memory pressure problem and for ease of comparison with the algorithm we will propose in the later sections.
Instead of the greedy approach proposed by Chen et al., which uses some heuristic to find good results in seconds, we formulate our first optimization problem as {\em checkpoint selection problem} based on Algorithm 2 and propose an $O(n^3)$ algorithm for finding the optimal checkpoint subset using dynamic programming.
Even more, by tracing the memory usage reported from PyTorch and comparing it with our simulated results, we revise Algorithm 2 into Algorithm~\ref{alg:train-early-free} using the theoretical analysis of model training we have mentioned at the beginning and denote the new problem as {\em dynamic checkpoint selection}. We formulate an accurate objective function based on the tracing and propose an $O(n)$-time algorithm for finding the optimal checkpoint subset using dynamic programming.
%Through extensive experiments, we conclude that Algorithm~\ref{alg:train-early-free} can precisely describe the behavior of the state-of-the-art deep learning platform PyTorch.
Through extensive experiments, we conclude that our algorithm can precisely describe the behavior of the state-of-the-art deep learning platform PyTorch.

\subsection{Analysis of Backward Propagation}

We describe the computation of a neural network model abstractly.
The model has $n$ weights $w_1, \ldots, w_n$, an input $x$, and the ground truth $G$.
The training for input $x$ consists of a {\em forward phase} and a {\em backward phase}, which will update the weights.
During an epoch, the training uses a set of different inputs to train the weights.
The training repeats the epoch until the weights converge.

The forward phase computes data $d_i$ with data $d_{i-1}$ and weight $w_i$ with a function $f_i$, for $i$ from $1$ to $n$. 
Note that we define $d_0$ to be the input $x$ to ease the notation in Equation~\ref{eq:di}. 

\begin{equation}
d_i = f_i(d_{i-1}, w_i) \label{eq:di}
\end{equation}

Then we start the {\em backward phase} that updates weights according to a loss function.
We compare the final output from Equation~\ref{eq:di} ($d_n$) with the ground truth $G$ corresponding to input $x$, and compute a loss function $L(d_n, G)$.
Now we compute the gradient $g_i$ of $w_i$ so that we can update it to minimize the loss function.
Since the loss is a function of the ground truth $G$ and $d_n$, and $d_n$ is a function of $d_{n-1}$ and $w_n$, and so on, we can compute the gradient $g_i$ of $w_i$ as a product in Equation~\ref{eq:chain} by the chain rule.

\begin{equation}
g_i = \frac{\partial d_i}{\partial w_i} (\prod_{j=i+1,n} \frac{\partial d_j}{\partial d_{j-1}}) \frac{\partial L}{\partial d_{n}} \label{eq:chain}
\end{equation}

Note that in practice we do not need to compute $g_i$ as the product in Equation~\ref{eq:chain}.
Instead, let $r(i)$ be a part of the product in Equation~\ref{eq:chain} then we have the following recursion.

\begin{eqnarray}
r_n & = & \frac{\partial d_n}{\partial d_{n-1}} \frac{\partial L}{\partial d_{n}}  \label{eq:rn} \\
r_i & = & \frac{\partial d_i}{\partial d_{i-1}} r_{i+1}, 1 \leq i \leq n - 1 \label{eq:ri}
\end{eqnarray}

Now we can rewrite Equation~\ref{eq:chain} as Equation~\ref{eq:gi}.

\begin{equation}
g_i = \frac{\partial d_i}{\partial w_i} r_{i+1} \label{eq:gi}
\end{equation}

We now compute the gradients and update the weights {\em backward}.
We first compute $g_n$ from Equation~\ref{eq:rn} and \ref{eq:gi}, and update weight $w_n$.
We then compute the rest of the gradient $g_i$ ($1 \leq i \leq n - 1$) from Equation~\ref{eq:ri} and \ref{eq:gi}, then update weight $w_i$ with gradient $g_i$.  
This is called {\em backward propagation} in the literature.
Algorithm~\ref{alg:train} gives the pseudocode for the training.

\begin{algorithm}
\begin{algorithmic}
\caption{Neural Network Training}
\label{alg:train}
%\Comment{Forward phase}
\Require Weights $w_1, \ldots, w_n$, input $x$ ($d_0$), and the ground truth $G$.
\Ensure Updated $w_1, \ldots, w_n$.\\
        \State {Allocate memory for $d_i$, $1 \leq i \leq n$}
\For{$i \gets 1$ to $n$}        \Comment{Forward Phase}         
        \State {Compute $d_i$ with $d_{i-1}$ and $w_i$ (Equation~\ref{eq:di})}
\EndFor
\\ 
\For{$i \gets n$ to $1$}     \Comment{Backward Phase}
    \State {Compute $r_{i}$ (Equation~\ref{eq:rn} or \ref{eq:ri})}
    \State {Compute $g_i$ (Equation~\ref{eq:gi})}
    \State{Update $w_i$ with $g_i$} \Comment{Update the weight}
\EndFor 
\end{algorithmic}
\end{algorithm}

Most of the current deep-learning platforms allocate memory for all data before the computation starts to ensure memory is available during the entire training.
For example, Algorithm~\ref{alg:train} will allocate memory for all $d_i$, $1 \leq i \leq n$, so that the backward phase can proceed without allocating any more memory.
However, this allocation causes severe and unnecessary memory pressure, since the backward phase proceeds in layers and does not need all $d_i$ simultaneously.

Chen et al.~\cite{chen2016training} proposed a {\em checkpoint} technique that reduces memory pressure.
The checkpoint technique computes all the data $d_i$ during the forward phase just as in Algorithm~\ref{alg:train}, but only keeps a subset (called checkpoints) in memory. 
During the backward phase, we can {\em recompute} the data $d_i$ between two checkpoints so that the backward phase can proceed.
For ease of explanation, we will denote these non-checkpoints between two checkpoints as a {\em segment}.
It is easy to see that once we complete the backward phase on a segment, we can free it and allocate memory for the next segment.
As a result, the maximum memory usage is the sum of the checkpoints, since they are always in memory during the backward phase, and the maximum amount of memory of a segment since only one of them will appear in memory at any given time.
That is, two data $d_i$'s from different segments will not be in memory simultaneously, and we can reduce the peak memory requirement.
The pseudocode of the checkpoint algorithm is in Algorithm~\ref{alg:train-checkpoint}.

\begin{algorithm}[h!tb]
\begin{algorithmic}
\caption{Neural Network Training with Checkpoint}
\label{alg:train-checkpoint}
%\Comment{Forward phase}
\Require Weights $w_1, \ldots, w_n$, input $x$ ($d_0$), and the ground truth $G$.
\Ensure Updated $w_1, \ldots, w_n$.\\
\For{$i \gets 1$ to $n$}        \Comment{Forward Phase}         
        \State {Allocate memory for $d_i$}
        \State {Compute $d_i$ with $d_{i-1}$ and $w_i$ (Equation~\ref{eq:di})}
        \If {$d_{i-1}$ is not a checkpoint}
            \State {Free the memory of $d_{i-1}$} 
        \EndIf
\EndFor
\For{$i \gets n$ to $1$}     \Comment{Backward Phase}
    \If {$d_i$ is a checkpoint}
        \State {Free the memory of the segment after $d_i$}
        \State {Let $d_h$ be the previous checkpoint, and allocate $d_{h+1}, \ldots, d_{i-1}$}
        \State {Recompute $d_{h+1}, \ldots, d_{i-1}$ (Equation~\ref{eq:di})}
    \EndIf
    \State {Compute $r_i$ (Equation \ref{eq:ri})}
    \State {Compute $g_i$ (Equation~\ref{eq:gi})}
    \State{Update $w_i$ with $g_i$} \Comment{Update the weight}
\EndFor 
\end{algorithmic}
\end{algorithm}

One immediate question one needs to answer is which subset from the data set $\{d_1, \ldots, d_n\}$ should be checkpoints.
There is a trade-off between the amount of memory for checkpoints and the maximum-sized segment, i.e. the segment with the maximum sum of data memory.
We can choose more checkpoints to reduce the memory of the maximum-sized segment, but the memory for checkpoints will increase.
On the contrary, choosing fewer checkpoints will increase the memory of the maximum-sized segment.
Therefore, it is essential to choose the correct checkpoints to balance the memory of the maximum-sized segment and the checkpoints.

%% file: 2_related.tex
\section{Related Work}\label{sec:related}

In this section, we describe some works related to the checkpoint technique in the previous section for reducing the peak memory requirement.

In Deep Neural Networks training, the memory of data for backward propagation has outweighed that of model parameters~\cite{sohoni2022lowmemory}. 
This causes severe memory pressure in Algorithm~\ref{alg:train}. 
Chen et al. introduced checkpoints (Algorithm~\ref{alg:train-checkpoint}) to reduce the memory pressure by recomputing and gave a very simple estimate on the number of checkpoints required~\cite{chen2016training}.
After that several efforts tried to find the optimal set of checkpoints to reduce the peak memory requirement~\cite{feng2021optimal}. 
\comment{
inally, we list a few applications where the checkpoint technique is used in combination with some other techniques to solve different kinds of problems.
}

\subsection{Trading computation for memory}

The question about the trade-off between allocating more memory for either checkpoints or the maximum-sized segment at the end of the previous section corresponds to the general methodology known as {\em trading computation for memory}.
Chen et al.~\cite{chen2016training} showed that one can train a $n$-layer linear chain feedforward model in $O(\sqrt{n})$ memory at the cost of an additional forward pass in the backward phase during training.
\comment{We can verify it by dividing the model into $\sqrt{n}$ number of equal-size segments and by only keeping the output (the checkpoints) of these segments during training. }
The memory cost to train the model is $O(n/\sqrt{n})+O(\sqrt{n}) = O(\sqrt{n})$, which is optimal when the size of data of every layer are the same.
In practice the $\sqrt{n}$ estimate might not be an optimal solution since the amount of data in different layers will differ.
As a result, dividing the model into $\sqrt{n}$ number of equal-size segments does not guarantee the optimal results.
Nevertheless, the simple technique still reduces the peak memory requirement.

Modern deep-learning platforms do support checkpoints.
For example, PyTorch~\cite{paszke2019pytorch} provides the tool API \texttt{torch.utils.checkpoint\_sequential} to implement Algorithm~\ref{alg:train-checkpoint} for sequential models and \texttt{torch.utils.checkpoint} to specify segments by adding checkpoints.
However, these functions are user-dependent, which means that its users have to find a set of checkpoints and divide their model into segments on their own accordingly to facilitate the tool API. To get the optimal result for all cases on linear models, we need to look for algorithms that work for models with non-uniform costs.

\subsection{Finding checkpoints within a given budget automatically}

To find the set of checkpoints, Chen et al. proposed another algorithm~\cite{chen2016training} in their work to generate a {\em memory allocation plan} (the placement of the checkpoints) within a given {\em budget} (the restriction on the memory of the maximum-sized segment). 
The algorithm can be seen as the {\em preprocessing step}~\cite{feng2021optimal} before the execution of Algorithm~\ref{alg:train-checkpoint}. 
In addition, it can be applied to general computation graphs without the assumption of uniform cost of layer data.
In short, it works by running a heuristic search over all possible budgets, and for each given budget it does greedy allocation to get the exact memory for each memory allocation plan.
\citet{herrmann2019optimal} also proposed a method to find the checkpoints under a given memory budget but focused on minimizing the model training time.
Their method is based on a dynamic programming algorithm of $O(mn^3)$ time complexity, where $m$ is the memory limit and $n$ is the number of layers.
In contrast, this paper focuses on finding the optimal checkpoints with minimal peak memory usage, and we propose a linear time algorithm to solve the checkpoint selection problem.

\comment{While the algorithm relies on approximate memory estimation for faster speed, it still takes several seconds to generate the plan, as reported in the supplement section of their work. 
Even worse, the solution is not optimal due to the approximate estimation. 
Instead, our work will improve these two parts in the later sections and achieve a linear algorithm with precise memory estimation.}

\subsection{Finding the optimal checkpoints for arbitrary DNN model}

To find the optimal set of checkpoints to reduce peak memory, the scheme proposed by Feng et al.~\cite{feng2021optimal} is the first work that is applicable for arbitrary computation graphs (ACGs), as they did not pose any assumption on the computation graph of DNN models. They can achieve maximum memory cut-offs at the cost of moderate time overhead (\textasciitilde 30\%-50\%).

To provide an overview of their algorithm, they denoted the computation graph of a DNN model as an acyclic-directed graph where the vertices are the intermediate tensors and the edges are the operations of the DNN model~\cite{feng2021optimal}. 
After the computation graph of a DNN model is built, they run their algorithm to divide the computation graph into {\em independent segments} (ISs), which means that there is no data dependency between the non-checkpoint vertexes in the subgraph and the other part of the graph. While the context is different, they use the same objective function for the peak memory requirement as the one defined by Chen et al, i.e. the memory of each segment is the sum of the memory of the non-checkpoint vertexes within the segment.

\subsection{Summary}

Table~\ref{tbl:comparison} summarizes checkpoint models and algorithms for linear neural networks.
%
%The first half of the table describes the memory model of Algorithm~\ref{alg:train-checkpoint}.
%This paper improves the execution time to find the minimum memory for this memory model from $O(n^3 \log n)$ to $O(n^3)$ (Section~\ref{sec:partition}).
%The algorithm is suitable for deep learning frameworks that do not optimize memory usage.
The upper half of Table~\ref{tbl:comparison} focuses on the memory usage model proposed by Chen et. al.~\cite{chen2016training}.
In this paper, we improve the execution time for finding the minimum memory usage from $O(n^3 \log n)$ to $O(n^3)$.

%In the second half of the table, as we will see in Section~\ref{sec:dynamic_checkpoint}, the memory model of Algorithm~\ref{alg:train-checkpoint} is inconsistent with the PyTorch implementation, which may lead to non-optimal memory usage.
%PyTorch implementation will not keep all the checkpoints in memory during the entire backward phase.
%Instead, it will release those checkpoints when they will not be used afterward.
%Therefore we derive a more accurate memory usage model, including the output gradient, and design an optimal linear time algorithm that ensures optimal memory utilization in PyTorch (Section~\ref{sec:dynamic_checkpoint}).
%Note that this paper focuses on finding the minimum memory footprint and does not consider the trade-off between memory and computation time.
The lower half of Table~\ref{tbl:comparison} focuses on the memory usage model derived from PyTorch's implementation.
The memory model of Algorithm~\ref{alg:train-checkpoint} is inconsistent with that of PyTorch's implementation.
In this paper, we derive the memory usage model that fits the memory usage reported by PyTorch, as will be demonstrated in Section~\ref{sec:experiment}.
Based on this accurate memory usage model, we design a linear time algorithm that ensures optimal memory utilization (Section~\ref{sec:dynamic_checkpoint}).
Note that this paper focuses on finding the minimum memory footprint and does not consider the trade-off between memory and computation time.

\newpage
\begin{table} 
\begin{center}
\begin{tabular}{|l|c|c|c|c|}
\hline
model / works & \cite{chen2016training} & \cite{feng2021optimal} & \cite{herrmann2019optimal} & this paper \\ \hline \hline
checkpoint & \Checkmark & \Checkmark & \Checkmark & \Checkmark \\ \hline 
layers of different sizes & & \Checkmark& \Checkmark& \Checkmark \\ \hline
memory model of Algorithm~\ref{alg:train-checkpoint} & \Checkmark & \Checkmark& & \Checkmark \\ \hline
the execution time of Algorithm~\ref{alg:train-checkpoint} & & $O(n^3 \log n)$ & & $O(n^3)$ \\
\hline
\hline
%consider memory usage optimization (e.g. Pytorch) & & & \Checkmark & \Checkmark\\ \hline
%consider output gradient memory in Pytorch & & & \Checkmark & \Checkmark\\ \hline
consider memory usage optimization (e.g. PyTorch) & & & \Checkmark & \Checkmark\\ \hline
consider output gradient memory in PyTorch & & & \Checkmark & \Checkmark\\ \hline
trade-off between memory and execution time & & & \Checkmark & \\ \hline
%optimal dynamic programming finding the minimum memory & & &  & \Checkmark \\ \hline
find the minimum memory usage & & &  & \Checkmark \\ \hline
\end{tabular}
\end{center}
\caption{The comparison of optimization on linear neural networks} \label{tbl:comparison}
\end{table}

\comment{
\subsection{Beyond finding the optimal peak memory using the checkpoint technique}

Many other schemes using the checkpoints have been proposed to solve different problems.

Kirisame et al.~\cite{kirisame2021dynamic} proposed a scheme called {\em Dynamic Tensor Rematerialization}, which is a greedy online algorithm that can checkpoint any DNN model and achieve comparable performance with the current checkpoint techniques. In the simulation, they can achieve $\Omega(\sqrt{N})$ memory budget with only $O(N)$ tensor operations for a $n$-layer linear feedforward network.
}

%% file: 3_algorithm.tex
\section{Checkpoint Selection Problem}\label{sec:partition}

\subsection{Problem Definition}

We consider a sequence of $n+1$ data $(d_0, \dots, d_n)$, where $d_0$ is the input and $d_n$ is the output, as in Section~\ref{sec:introduction}.
For ease of explanation, $d_i$ denotes both the data and the size of the data.
Since $d_0$ is the input and it needs to be in memory, we choose it as a checkpoint.
In addition, the training generates the output $d_n$ after the forward phase and immediately uses it at the beginning of the backward phase, so we cannot save any memory by {\em not} choosing it as a checkpoint.
As a result, without loss of generality, we assume both $d_0$ and $d_n$ are checkpoints in this paper.

\begin{figure}[h!tb]
\vspace*{5mm}
\begin{center}
\begin{tikzpicture}
\draw[fill=black] (0,0) circle (0.3) node [black,yshift=-0.8cm] {$d_0$};
\draw[fill=none] (1,0) circle (0.3) node [black,yshift=-0.8cm] {$d_1$};
\draw[fill=black] (2,0) circle (0.3) node [black,yshift=-0.8cm] {$d_2$};
\draw[fill=none] (3,0) circle (0.3) node [black,yshift=-0.8cm] {$d_3$};
\draw[fill=none] (4,0) circle (0.3) node [black,yshift=-0.8cm] {$d_4$};
\draw[fill=black] (5,0) circle (0.3) node [black,yshift=-0.8cm] {$d_5$};
\end{tikzpicture}
\end{center}
\vspace*{-10mm}
\caption{An example of three checkpoints (in black) and two segments: \{d1\} and \{d3, d4\} (in white)} \label{fig:layers}
\end{figure}
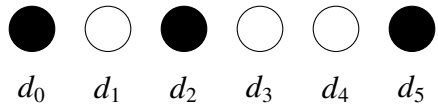

Now we choose $m$ additional checkpoints to save memory by recomputing non-checkpoints during the backward phase, as in Algorithm~\ref{alg:train-checkpoint}.
Recall that the data between two checkpoints becomes a segment.
Now the set $C$ of $m + 2$ checkpoints (including $d_0$ and $d_n$)  will partition the sequence $(d_0, \dots, d_n)$ into $(s_1, \ldots, s_{m+1})$ segments, where each segment is the set of data between two checkpoints.
Note that two checkpoints may be adjacent so that a segment may be empty.
Figure~\ref{fig:layers} shows the partitioning of a model into three checkpoints and two segments.

From the previous discussion, we consider an objective function consisting of the {\em checkpoint memory} and the {\em segment memory}.
The {\em checkpoint memory} is the sum of the memory of all checkpoints, i.e., $\sum_{i \in C} d_i$.
The {\em segment memory} is the maximum of the sum of memory in every segment, i.e., $\max_i {\sum_{j \in s_i} d_j}$.
The goal is to find a checkpoint subset $C$ which minimizes the sum of checkpoint memory and segment memory (Equation~\ref{eq:cost}).
%The goal is to find a checkpoint subset $C$ that minimizes the sum of checkpoint memory and segment memory (Equation~\ref{eq:cost}).
We will denote this problem as the {\em checkpoint selection problem}.

\begin{equation}
\sum_{i \in C} d_i + \max_{1 \leq i \leq m+1}({\sum_{j \in s_i} d_j}) \label{eq:cost}
\end{equation}

\subsection{Dynamic Programming}

In this section, we solve the checkpoint selection problem by an algorithm based on dynamic programming.

The algorithm consists of two steps:
\begin{enumerate}
    \item For each possible segment memory $s$, find the minimal checkpoint memory $c_s$ such that the segment memory does not exceed $s$.
    \item Find the minimum of $s + c_s$.
\end{enumerate}
The implementation of step 2 is a simple algorithm iterating over all possible $s + c_s$.
In the next paragraph, we will discuss how to find $c_s$ for all $s$ in step 1.

Step 1 of the algorithm is implemented using dynamic programming as follows.
Define a function $M(i, s)$ as the minimum checkpoint memory for the sequence $(d_1, \ldots, d_i)$ such that the segment memory does not exceed $s$.
Here $c_s$ equals $M(n, s)$.
Now we derive the recursion for $M$.
%We first consider a more accessible version of the checkpoint selection problem in which we know the maximum segment memory.
%Let $s$ be the maximum segment memory allowed; how do we compute the minimum checkpoint memory?
%We derive dynamic programming that solves this problem for a segmented memory $s$.
%We define $M(i, s)$ as the minimum checkpoint memory for the sequence $(d_1, \ldots, d_i)$, subject to the condition that the segment memory is no more than $s$.
%Each $M(i, s)$ is the sum of the checkpoint $d_j$, closest to $d_i$, and the minimum checkpoint memory between $(d_1, \ldots, d_{j-1})$.
Each $M(i, s)$ is the sum of the checkpoint $d_j$, closest to $d_i$, and the minimum checkpoint memory between $(d_0, \ldots, d_{j-1})$.
%We now consider the possibility of choosing $d_j$ as a checkpoint.
Since the segment memory does not exceed $s$, the term $j$ must satisfy Inequality~\ref{ineq:j}.
%From the assumption that the segment memory is bounded by $s$, we have Inequality~\ref{ineq:j}.
%Note that we only consider $j$ so that the segment cost from $d_{j+1}$ to $d_{i}$ is no more than the segment cost $s$ we assumed.  
\begin{equation}
\sum_{k = j + 1}^i d_k \leq s \label{ineq:j}
\end{equation}
Therefore, the term $M(i, s)$ is the minimum of $d_j + M(j - 1, s)$ for all $j$ satisfying Equation~\ref{ineq:j}.
To express the recursion as a formula, let $l(i)$ be the minimum $j$ that satisfies Inequality~\ref{ineq:j}.
The recursion for $M$ is given in Equation~\ref{eq:T}.
%Now we consider all possible $j$'s that satisfy Inequality~\ref{ineq:j}.
%Let $l(i)$ be the minimum $j$ that satisfies Inequality~\ref{ineq:j}.
%Note that $l(i)$ is a non-decreasing function of $i$.
%We now derive the recursion for $M(i, s)$.

\begin{equation}
M(i, s) = \min_{l(i) \leq j \leq i} (d_j + M(j - 1, s)) \label{eq:T}
\end{equation}

\begin{algorithm}[h!tb]
\begin{algorithmic}
\caption{Checkpoint Selection}
\label{alg:checkpoint-selection}
%\Comment{Forward phase}
\Require data sizes $d_0, \ldots, d_n$.
\Ensure The minimum cost of checkpoint selection
\For{all segment cost $s$}
    \State{Compute each $l(i)$ for $1 \leq i \leq n$}
    \State{$M(0, s) = d_0$}
%   \State{$M(s, 0) = d_0$}
    \State{$Q = \{0\}$}
    \For{$i \gets 1$ to $n$}
%        \State {Remove those $q_h$ from the head of $Q$ such that $l(i) > j$.}
        \State {Remove those $q_h$ from the head of $Q$ such that $l(i) > q_h$.}
        \State {Set $M(i, s)$ to be $d_q + M(q - 1, s)$, where $q$ is the head of $Q$.}
        \State {Remove those $q_t$ from the tail of $Q$ such that $d_{q_t} + M(q_t - 1, s) > d_i + M(i - 1, s)$.}
        \State {Add $i$ to the tail of $Q$.}
    \EndFor
%    \State{$Q = \emptyset$}
%    \For{$i \gets 1$ to $n$}
%        \State {Remove from the end of $Q$ those element greater than $d_i + M(i - 1, s)$}
%        \State {Add $d_i + M(i - 1, s)$ to the tail of $Q$}
%        \State {Set $M(i, s)$ to be the first element of $Q$}
%    \EndFor        
\EndFor
\State{return $\min_{s} (s + M(n, s))$} \Comment{the answer}
\end{algorithmic}
\end{algorithm}

Algorithm~\ref{alg:checkpoint-selection} is the implementation details of the algorithm that solves the checkpoint selection problem.
%We can now use Equation~\ref{eq:T} to solve the checkpoint selection problem as in Algorithm~\ref{alg:checkpoint-selection}.
%First, we enumerate all possible segment cost $s$, then we use Equation~\ref{eq:T} to solve $M(n, s)$ for this given $s$.
%Finally, we find the minimum among all $M(n, s)$ for all $s$, which is the answer to our checkpoint selection problem.
%We will analyze the time complexity of using Equation~\ref{eq:T} to solve $M(n, s)$ for a given $s$, and the number of segment costs $s$ we need to enumerate in the following two paragraphs.
%The head of $Q$, say $q$, must be the one with the smallest $d_q + M(q - 1, s)$ since each time we add an element $u$ to $Q$, we remove those $v$ from the end of $Q$ such that $d_u + M(u - 1, s) < d_v + M(v - 1, s)$.
%Therefore, we have $M(i, s) = d_q + M(q - 1, s)$.
% We use a queue $Q$, which contains the $j$'s in Equation~\ref{eq:T}, to implement the recursion of $M$.
We claim that, in order to prove the correctness of our algorithm,
\begin{itemize}
    \item the queue $Q$ contains the $j$'s achieving the minimum in Equation~\ref{eq:T}, and
%    \item the head of $Q$ is such a $j$.
    \item the head of $Q$ is $\displaystyle\argmin_{l(i)\leq j\leq i}\,(d_j + M(j - 1, s))$.
\end{itemize}

The latter is valid, since in the $i$-th iteration, we remove those $j$ from the tail of $Q$ such that $d_i + M(i - 1, s) < d_j + M(j - 1, s)$.
Therefore, the head of $Q$, say $q$, is the element with the smallest $d_q + M(q - 1, s)$ in $Q$.

To prove the former, we ensure that in the $i$-th iteration, all $j$'s with $j < i$ have entered $Q$, and only the ones that do not achieve the minimum in Equation~\ref{eq:T} are removed from $Q$.
All $j$'s with $j < i$ have entered $Q$ since we add each $j$ to the tail of $Q$ before the end of the $j$-th iteration.
If $u < i$ is removed from $Q$ because some $v < i$ induces $u < l(v)$, then since $l$ is monotonically increasing, we have $l(v) < l(i)$, implying that $u < l(i)$.
Therefore, the term $u$ is not a valid $j$ in Equation~\ref{eq:T}.
If $u < i$ is removed from $Q$ because some $v < i$ induces $d_u + d(u - 1, s) > d_v + d(v - 1, s)$, then whenever $u$ is a valid $j$ in Equation~\ref{eq:T}, we can always find another $j$, namely $v$, with a smaller $d_j + M(j - 1, s)$.
Therefore, the term $u$ is not the $j$ achieving the minimum in Equation~\ref{eq:T}.

We analyze the time complexity of Algorithm~\ref{alg:checkpoint-selection} in the following paragraphs.
To analyze the time complexity of completing the recursion $M(i, s)$ for a fixed $s$, three operations are put into consideration: computing each $l(i)$ with $1 \leq i\leq n$, assigning values to each $M(i, s)$, and maintaining the queue $Q$.
%To analyze the time complexity of completing the recursion $M(i, s)$ for a fixed $s$, two operations are put into consideration: assigning values to each $M(i, s)$ and maintaining the queue $Q$.
\begin{itemize}
\item Computing each $l(i)$ takes $O(n)$ time by maintaining the sliding window that starts at $l(i)$ and ends at $i$ over the sequence $(1, 2,\dots, n)$.
This is because each element in the sequence $(0, 1,\dots, n)$ enters and exits the window only once, and each insertion and deletion takes $O(1)$ time.
\item In the $i$-th iteration for each segment memory usage $s$, it takes $O(1)$ time to find the head of $Q$, say $q$, and assign $d_q + M(q - 1, s)$ to each $M(i, s)$.
Therefore, it takes $O(n)$ time to assign values to each $M(i, s)$ for a fixed $s$.
%\item For each segment memory usage $s$, each number from $0$ to $n$ enters and leaves the queue $Q$ once, and therefore it takes $O(n)$ time to maintain the queue $Q$.
\item For each segment memory usage $s$, each number from $0$ to $n$ enters and leaves the queue $Q$ once, resulting in $O(n)$ insertions and deletions.
Since each insertion and deletion occurs only at the head and tail of $Q$, both operations have $O(1)$ time complexity.
Therefore, the maintenance of $Q$ takes $O(n)$ time.
\end{itemize}
As a result, it takes $O(n)$ time to complete the recursion of $M(i, s)$ for a fixed $s$.

Next, we find the number of different segment memory usages.
%The next question is determining the segment cost $s$ for $M$ in Equation~\ref{ineq:j} and \ref{eq:T}.
We cannot iterate each value between $1$ and $\sum_{i = 1}^n d_i$ since it leads to a pseudo-polynomial time algorithm.
%We cannot test all possible values below $\sum_{i = 1}^n d_i$ since it leads to a pseudo-polynomial time algorithm, not a polynomial-time algorithm.
In fact, there are $O(n^2)$ possible segment memory usages, obtained by enumerating the start and end points of the segments.
%Instead, we observe only $O(n^2)$ possible segments since every segment has a starting and end point.

As a result, there are $O(n^2)$ different segment memory usages, each taking $O(n)$ time to compute.
%Therefore, Theorem~\ref{thm:checkpoint-selection-problem} follows.
Therefore, we obtain the following theorem.
%As a result, we only need to consider $O(n^2)$ different $s$'s in Equation~\ref{ineq:j} and \ref{eq:T}.
%We also know that the time complexity for computing $T$ with a given $s$ is $O(n)$ from the previous discussion; therefore, we conclude that the total time for solving the checkpoint selection problem is $O(n^3)$.

%After knowing all $M$ function values, we can solve the checkpoint selection problem by finding the $s$ that minimizes Equation~\ref{eq:sol}, where $s$ is the segment memory and $M(n, s)$ is the checkpoint memory.

%\begin{equation}
%\min_{s} (s + M(n, s) \label{eq:sol})
%\end{equation}

\begin{theorem}
\label{thm:checkpoint-selection-problem}
The checkpoint selection problem is $O(n^3)$-time solvable, where $n$ is the number of neural network layers.
%A dynamic program solves the checkpoint selection problem in $O(n^3)$ time, where $n$ is the number the neural network layers.
\end{theorem}

%% file: 4_dynamic_checkpoint.tex
\section{PyTorch Implementation}\label{sec:dynamic_checkpoint}

We use Algorithm~\ref{alg:train-checkpoint} to estimate the memory usage and compare it with the result from PyTorch, a state-of-the-art deep learning platform that supports checkpoints. 
However, the memory usage reported by PyTorch does not match the simulated results from Algorithm~\ref{alg:train-checkpoint}.
After tracing the implementation of PyTorch, we realized the following.

\begin{enumerate}
\item First, PyTorch does {\em not} keep all checkpoints in memory all the time, as Equation~\ref{eq:cost} indicates.  
We observe that PyTorch released the memory of $d_i$ (checkpoint or not) when it is no longer used.
That is, it does not release the memory in batch, as algorithm~\ref{alg:train-checkpoint} suggests.
\item Algorithm~\ref{alg:train-checkpoint} does not consider the memory of $r_i$ (Equation~\ref{eq:ri}).
The size of this {\em output gradient} is the same as $d_i$, which is quite significant.
In addition, PyTorch will allocate a data buffer to fit the maximum output gradient $d_i$ within a segment.
As a result, the size of maximum $d_i$ should be counted {\em twice} for an accurate estimate of the memory requirement of PyTorch. 
\end{enumerate}

We now describe a more memory-saving checkpoint mechanism. 
We conjectured that PyTorch uses this mechanism, instead of Algorithm~\ref{alg:train-checkpoint},
and we verified this conjecture by experiments in Section~\ref{sec:experiment}.
Therefore, this is a more accurate description of the PyTorch implementation.
The pseudocode of this memory management algorithm is in Algorithm~\ref{alg:train-early-free}.

\begin{algorithm}[h!t]
\begin{algorithmic}
\caption{Neural Network Training with Checkpoints as Done by PyTorch}
\label{alg:train-early-free}
%\Comment{Forward phase}
\Require Weights $w_1, \ldots, w_n$, input $x$ ($d_0$), and the ground truth $G$.
\Ensure Updated $w_1, \ldots, w_n$.\\

\For{$i \gets 1$ to $n$}        \Comment{Forward Phase}     
    \State {Allocate memory for $d_i$}  
    \State {Compute $d_i$ with $d_{i-1}$ and $w_i$}
    \If {$d_{i-1}$ is not a checkpoint}
        \State {Free the memory of $d_{i-1}$} 
    \EndIf
\EndFor
\For{$i \gets n$ to $1$}     \Comment{Backward Phase}
    \If {$d_i$ is a checkpoint}
        \State{Let $d_h$ be the previous checkpoint. Allocate and recompute $d_{h+1}, \ldots, d_{i-1}$}
%        \State{Allocate and recompute $d_{h+1}, \ldots, d_{i-1}$}
        \State{Release the buffer for the previous output gradients} 
        \State{Allocate a new buffer $r$ of size  $\max(d_{h}, \ldots, d_{i-1})$} 
        \Comment{output gradient}
    \EndIf
    \State {Compute $r_i$ (Equation \ref{eq:ri})}
    \State {Compute $g_i$ (Equation~\ref{eq:gi})}
    \State{Update $w_i$ with $g_i$} \Comment{Update the weight}
    \State{Release the memory $d_i$} \Comment{as PyTorch implementation}
\EndFor 
\end{algorithmic}
\end{algorithm}

The new mechanism will release the memory of $d_i$ as soon as we do not need them.
The forward phase is the same as the previous mechanism; we only keep checkpoints in memory.
During the backward phase, we will have two cases for the index $i$ in Algorithm~\ref{alg:train-early-free}.

\begin{itemize}
\item If $d_i$ is a checkpoint, then we will first find the previous checkpoint $d_{h}$ and recompute all data in the segment between $d_{h+1}$ and $d_{i-1}$, where $d_h$ is the previous checkpoint.
\item We then free the memory of the output gradient of the previous segment.
\item We then allocate a new buffer for the output gradient of this segment, of size $\max(d_{h}, \ldots, d_{i-1})$.
\item We then update the weight $w_i$ as in Algorithm~\ref{alg:train-checkpoint}.
Finally, just as in the PyTorch implementation, we free the memory of $d_{i}$ since we no longer need it.
\end{itemize}

With these observations, we derive a more accurate memory consumption model and verify it with experiments in Section~\ref{sec:prediction}.
Now we focus on the theoretical aspects of finding the set of checkpoints to minimize memory use for this new model.

One can think of this new checkpoint mechanism as it shrinks the current segment until it reaches the next checkpoint, and then it recomputes the next segment and starts shrinking it. 
Also, it always keeps a buffer of the size of the maximum among the current segment and its left checkpoint as the output gradient.
Let $C$ be the set of checkpoints, $d_i \in C$, and $d_h$ is the checkpoint before $d_i$ in $C$. 
The memory usage for the $i$-th backward phase $m(i)$ is in Equation~\ref{eq:memory}, where $C(i)$ is the set of checkpoints with an index no larger than $i$.
Note that we only need to define $m(i)$ for those $i$'s where $d_i \in C$ to find the maximum memory usage since the memory usage from the $i$-th iteration to the $h-1$-th iteration always decreases, since Algorithm~\ref{alg:train-early-free} always free memory at the end of the iteration.  
For example, if $d_{{i'}}$ is a data between $d_h$ and $d_i$,  i.e., $h < i' < i$, then memory usage $m(i')$ is in Equation~\ref{eq:iprime}.
Note that PyTorch implementation (Algorithm~\ref{alg:train-early-free}) still maintains the output gradient buffer of size $\max_{h \leq k < i}(d_k)$ when the backward phase goes to $d_{i'}$.
It is obvious that $m(i') < m(i)$, so we only need to define $m(i)$ for those $i$'s where $d_i \in C$ to find the maximum.

\begin{eqnarray}
m(i) & = & \sum_{d \in C(i)} d + \sum_{k = h+1}^{i-1} d_k + \max_{h \leq k < i}(d_k) \label{eq:memory} \\
m(i') & = & \sum_{d \in C(h)} d + \sum_{k = h+1}^{i'} d_k + \max_{h \leq k < i}(d_k) \label{eq:iprime} 
\end{eqnarray}

For ease of notation, we define $s(h, i)$ as in Equation~\ref{eq:s}, so we can rewrite Equation~\ref{eq:memory} into \ref{eq:memory-simple}.

\begin{eqnarray}
s(h, i) & = & \sum_{k = h+1}^{i-1} d_k + \max_{h \leq k < i}(d_k) \label{eq:s} \\
m(i) & = & \sum_{d \in C(i)} d + s(h, i) \label{eq:memory-simple}
\end{eqnarray}

Given the $d_i$ for $1 \leq i \leq n$, the goal is to find the set of checkpoints $C$ which minimizes the maximum among all $i$ for Equation~\ref{eq:memory-simple}, for all $d_i$ in $C$.
We will use {\em dynamic checkpoint selection problem} to denote this optimization problem.
%We will use {\em dynamic checkpoint selection} to denote this optimization problem.
\subsection{Dynamic Programming}

We again use dynamic programming to solve the dynamic checkpoint selection problem.

Define a function $M(i)$, representing the minimum memory usage from layer $i$ to layer $n$, given that layer $i$ is the checkpoint.
Here, the term $M(1)$ is the answer to the dynamic checkpoint selection problem.
%We first define a memory usage function $M(i)$, which is the minimum memory usage {\em before} the step $i$ of the backward phase, considering data from $d_i$ to $d_n$, where $d_i$ is a checkpoint.
%
Now we derive the recursion of $M$.
For each term $M(i)$, $1\leq i < n$, consider the next checkpoint: $d_j$, $i < j < n$.
%Since $d_i$ is a checkpoint, we consider a possible next checkpoint, $d_j$, during the backward phase, for $i < j < n$.
There are two possibilities for minimum memory usage:
\begin{itemize}
\item The memory usage peaks when the backpropagation updates the weights between layer $j$ and layer $n$.
Therefore, the memory usage is $d_i + M(j)$.
\item The memory usage peaks when the backpropagation updates the weights between layer $i$ and layer $j$.
Therefore, the memory usage is $d_i + s(i, j) + d_j$.
\end{itemize}
%There are two possibilities for minimum memory usage during the backward phase.
%The first possibility happens before the $j$ step, so the memory usage is $d_i + M(j)$.
%That is, the memory usage peaks before the $j$-th iteration, so as far as $d_i$ to $d_n$ are concerned, the maximum is $d_i + M(j)$.
%The second possibility happens at the beginning of the $j$-th iteration, so the memory usage is $d_i + s(i,j) + d_j$.
As a result, we derive the recursion of $M$ as in Equation~\ref{eq:T-rec}.
%Since we do not know which $j$ will minimize the memory usage for an $i$, we enumerate all $j$ between $i$ and $n$ and find the one that minimizes the larger values of the two possibilities.
\begin{equation}
M(i) = \min_{i < j < n}(\max(d_i + M(j),  d_i + s(i,j) + d_j)) \label{eq:T-rec}
\end{equation}

%We rewrite Equation~\ref{eq:T-rec} into Equation~\ref{eq:T-rec-simple}.
%\begin{equation}
%M(i) = d_i + \min_{i < j < n}(\max(M(j),  s(i,j) + d_j)) \label{eq:T-rec-simple}
%\end{equation}

Here we discuss the time complexity of this recursion.
To find a single $M(i)$ using Equation~\ref{eq:T-rec}, we enumerate all $j$ between $i$ and $n$ and then find the one that minimizes the larger values between $d_i + M(j)$ and $d_i + s(i, j) + d_j$.
%We need to enumerate all $j$ between $i$ and $n$, compute $\max(M(j),  s(i,j) + d_j)$, and find their minimum.
A simple algorithm takes $O(n)$ time to compute each $M(i)$, and there are $n$ such $M(i)$ values to compute.
Therefore, the time for computing all $M(i)$'s is $O(n^2)$, if we precompute all $s(i,j)$, which also takes $O(n^2)$ time.
%A simple algorithm will take $O(n)$ time to compute a $M(i)$, and the total time for computing all $M(i)$'s will be $O(n^2)$ if we precompute all $s(i,j)$, which also takes $O(n^2)$ time.
See Algorithm~\ref{alg:dynamic-checkpoint-selection-square-time} for the implementation details.

\begin{algorithm}[h!tb]
\begin{algorithmic}
\caption{Dynamic Checkpoint Selection in $O(n^2)$ time}
\label{alg:dynamic-checkpoint-selection-square-time}
%\Comment{Forward phase}
\Require data sizes $d_0, \ldots, d_n$.
\Ensure The minimum cost of checkpoint selection
\State{$M(n)=d_n$}
\For{$i \gets n-1$ to $1$}
\State{$M(i)=\infty$}
\For{$j \gets i+1$ to $n$}
    \State{$M(i) = \min(M(i), \max(d_i + M(j),  d_i + s(i,j) + d_j))$}        
\EndFor
\EndFor
\State{return $M(1)$} \Comment{the answer}
\end{algorithmic}
\end{algorithm}

\begin{theorem}
\label{thm:dynamic-checkpoint-selection-square-time}
The dynamic checkpoint selection problem is $O(n^2)$-time solvable, where $n$ is the number of neural network layers.
%We can solve the dynamic checkpoint selection problem in $O(n^2)$ time, where $n$ is the number of layers.
\end{theorem}

%\comment{
\subsection{Linear Time}

%We now describe how to compute Equation~\ref{eq:T-rec-simple} in linear time of $n$.
%The first observation is that building a prefix sum table on $d_i$ will take $O(n)$ time.
%After we have the prefix sum table, we can compute any range sum of $d_i$, e.g., $\sum_{s=i+1}^{k} d_s$ of %Equation~\ref{eq:T-rec-simp}, in $O(1)$ time.
%We first observe that $M(i)$ is a decreasing function of $i$.

%\begin{lemma} \label{lm:T}
%$M(i)$ is a decreasing function of $i$.
%\end{lemma}
%\begin{proof}
%A direction observation from Equation~\ref{eq:T-rec-simple}, and the fact that $s(i,j) > s(i + 1,j)$.

%\begin{eqnarray*}
%\lefteqn{M(i)} \\
%& = & d_i + \min_{i < j < n}(\max(M(j),  s(i,j) + d_j)) \\
%& > & \min((\max(M(i+1), s(i, i+1) + d_{i+1}), \\ 
%& & \min_{i + 1 < j < n}(\max(M(j),  s(i,j) + d_j)) \\
%& > & \min(M(i+1), \min_{i + 1 < j < n}(\max(M(j),  s(i + 1,j) + d_j)) \\
%& > & \min(M(i+1), M(i+1)) \\
%& > & M(i+1)
%\end{eqnarray*}
%\end{proof}

We now describe how to compute Equation~\ref{eq:T-rec} in linear time of $n$.
First, we rewrite Equation~\ref{eq:T-rec} into Equation~\ref{eq:T-rec-simple}.
\begin{equation}
M(i) = d_i + \min_{i < j < n}(\max(M(j),  s(i,j) + d_j)) \label{eq:T-rec-simple}
\end{equation}
To ease the notation we define $U(i, j)$ as in Equation~\ref{eq:U} and simplify Equation~\ref{eq:T-rec-simple} into \ref{eq:TU}.

\begin{eqnarray}
U(i, j) & = & s(i,j) + d_j \label{eq:U} \\
M(i) & = & d_i + \min_{i < j < n}(\max(M(j), U(i, j))) \label{eq:TU}
\end{eqnarray}

The linear-time algorithm is derived from two observations:
\begin{itemize}
    \item The computation of $M(i)$ using Equation~\ref{eq:TU} can be reduced to finding the minimum of the larger values between an increasing function and a decreasing queue.
    \item The index that yields the minimum in Equation~\ref{eq:TU} when computing $M(i)$ is no greater than the index that yields the minimum in Equation~\ref{eq:TU} when computing $M(i + 1)$.
\end{itemize}

\subsubsection{The First Observation}
The function $U(i, j)$ is monotonic of $j$ for a given $i$.
%We observe that $U(i, j)$ is a monotonic function of $j$ for a given $i$.
\begin{lemma} \label{lm:U}
$U(i, j)$ is an increasing function of $j$ for a given $i$.
\end{lemma}
\begin{proof}
\begin{eqnarray}
\lefteqn{U(i, j + 1)} \\
& = & s(i,j + 1) + d_{j+1} \\
& = & (\sum_{k = i+1}^{j} d_k + \max_{i \leq k \leq j}(d_k)) + d_{j+1} \label{eq:U-(i,dot)}\\
& > & d_j + \sum_{k = i+1}^{j-1} d_k + \max_{i \leq k \leq j}(d_k) \\
& > & (\sum_{k = i+1}^{j-1} d_k + \max_{i \leq k < j}(d_k)) + d_j\\
& = & U(i, j)
\end{eqnarray}
\end{proof}

% We use a queue $Q$ to store the known $M(i)$ values.
% Whenever a $M(i)$, where $0 \leq i < n$, is found, apply the following procedure on $Q$:
% \begin{enumerate}
%    \item Locate the last entry $M(i^\star)$ within $Q$, if one exists.
%    \item If $M(i^\star) \leq M(i)$ or if $Q$ is empty, insert $M(i)$ into $Q$; otherwise, eliminate $M(i^\star)$ from $Q$ and call this procedure recursively.
% \end{enumerate}

We use a monotonic queue $Q$ to store a sorted subset of $M(i)$'s, for $i$ between $1$ and $n$.
We compute $M(i)$ for $i$ from $n$ down to $1$.
After computing $M(i)$, we remove those $M$'s from the head of $Q$ that are no less than $M(i)$, then add $M(i)$ to the head of $Q$.
%After we compute $M(i)$, we remove those $M$'s at the beginnin of $Q$ that are no less than $M(i)$, then add $M(i)$ at the beginning of $Q$.
As a result, we acquire Lemma~\ref{lm:monotonicQueue}.
%As a result, $Q$ is a sequence of {\em decreasing} $M(i)$ values with increasing $i$ indices.
%For ease of notation, we use $Q_i$ to denote the monotonic queue $Q$ after we added $M(i)$ to it.

% Equation~\ref{eq:TU} shows that when computing $M(i)$, we should consider all $M(j)$ values, where $i < j < n$.
% However, we observe that we only need to consider the $M(j)$ values in $Q$ during the computation of $M(i)$.

\begin{lemma} \label{lm:monotonicQueue}
$Q$ is a monotonically decreasing queue with respect to the index of function $M$.
\end{lemma}
\begin{proof}
\comment{
Assume for contradiction that $Q$ is not a monotonically decreasing queue with respect to the index of function $M$.
Then there exist two adjacent entries $M(i)$ and $M(j)$ in $Q$ such that
\begin{equation*}
    i < j \hspace{1cm}\text{and}\hspace{1cm} M(i) < M(j)
\end{equation*}
However, according to the procedure for maintaining $Q$, $M(j)$ is removed prior to inserting $M(i)$.
Thus, $M(j)$ is not in $Q$, a contradiction.}
The lemma follows from the construction of $Q$.
\end{proof}

For ease of notation, we use $Q_i$ to denote the monotonic queue $Q$ after we added $M(i)$ to it.
We observe that only those $M(j)$'s in $Q$ need to be considered when applying Equation~\ref{eq:TU} to compute $M(i)$.
%The first key idea of the linear time proof is that we only need to consider those $M(j)$'s in $Q$ for Equation~\ref{eq:T-rec-simple} when we compute $M(i)$.
Formally we have Equation~\ref{eq:T-rec-queue}.

\begin{equation}
M(i) = d_i + \min_{j \in Q_{i-1}}(\max(M(j), U(i, j))) = d_i + \min_{i < j < n}(\max(M(j), U(i, j))) \label{eq:T-rec-queue}
\end{equation}
%\begin{equation}
%M(i) = \min_{j \in Q_{i-1}}(\max(M(j), U(i, j))) = \min_{i < j < n}(\max(M(j), U(i, j))) \label{eq:T-rec-queue}
%\end{equation}

\begin{lemma} \label{lm:Q}
When we compute $M(i)$ with Equation~\ref{eq:TU}, we only need to consider those indices $j$'s in $Q_{i-1}$, i.e., we have Equation~\ref{eq:T-rec-queue}.
%When we compute $M(i)$ with Equation~\ref{eq:T-rec-simple}, we only need to consider those indices $j$'s in $Q_{i-1}$, i.e., we have Equation~\ref{eq:T-rec-queue}.
\end{lemma}
\begin{proof}
Consider $M(a)$ that removes $M(b)$ during the construction of $Q$.
By the construction of $Q$, we have $M(a) \leq M(b)$.
In addition, we have $a < b$ because we compute $M(i)$ for $i$ from $n$ to $1$.
This implies that $U(i, a) < U(i, b)$ since Lemma~\ref{lm:U} states that $U(i, j)$ is increasing for a fixed $i$.
%By the construction, we have $a < b$ because we compute $M(i)$ for $i$ from $n$ to $1$, so we have $M(a) \leq M(b)$.
%In addition, $U(i, a) < U(i, b)$ from Lemma~\ref{lm:U} because $a < b$.
The preceding two inequalities, $M(a) \leq M(b)$ and $U(i, a) < U(i, b)$, imply $\max(M(a), U(i, a)) \leq \max(M(b), U(i, b))$, which indicates that we can safely ignore the case $j=b$ when computing $M(i)$ using Equation~\ref{eq:TU}.
Therefore, the Lemma follows and Equation~\ref{eq:T-rec-queue} is valid.
%As a result, $\max(M(a), U(i, a)) \leq \max(M(b), U(i, b))$, and we can safely ignore $b$ in Equation~\ref{eq:TU}, thus the theorem follows and Equation~\ref{eq:T-rec-queue} is valid.
\end{proof}

\comment{
\begin{lemma} \label{lm:Q}
For each $x$ such that $0 \leq x < n$, if
\begin{equation}
i = \arg \min_{x < j < n}(\max(M(j), U(x, j)))
\end{equation}
then $M(i) \in Q$.
\end{lemma}
\begin{proof}
We prove Lemma~\ref{lm:Q} by showing that for each $x$ such that $0 \leq x < n$, if there exists an $i^*$ such that $M(i^*) \notin Q$, then
\begin{equation}
i^* \neq \arg\min_{x < j < n}(\max(M(j), U(x, j)))
\end{equation}

According to the procedure for maintaining $Q$, if $M(i^*) \notin Q$, then there exists an $i$, where $x < i < i^*$, such that $M(i) < M(i^*)$, implying that
\begin{equation} \label{eq:lmQTi}
M(i) < \max(M(i^*), U(x, i^*))
\end{equation}

By Lemma~\ref{lm:U} and $x < i < i^*$, we have $U(x, i) < U(x, i^*)$, implying that
\begin{equation} \label{eq:lmQUi}
U(x, i) < \max(M(i^*), U(x, i^*))
\end{equation}

By combining Equation~\ref{eq:lmQTi} and Equation~\ref{eq:lmQUi}, we obtain the following inequality:
\begin{equation}
\max(M(i), U(x, i)) < \max(M(i^*), U(x, i^*))
\end{equation}

Thus,
\begin{equation}
i^* \neq \arg \min_{x < j < n}(\max(M(j), U(x, j)))
\end{equation}
\end{proof}
}

\comment{
In addition, we observe that $Q$ is a monotonically decreasing queue with respect to the index of function $M$.
}

Combining Lemma~\ref{lm:U}, \ref{lm:monotonicQueue}, and \ref{lm:Q}, we obtain the first observation of our linear time algorithm: the value of $M(i)$ is the sum of $d_i$ and the minimum of the larger value between an increasing function $U(i,\cdot)$ and a decreasing queue $Q$ with respect to $j$, the index of $M$'s in $Q$.
%Combining Lemma~\ref{lm:U}, \ref{lm:monotonicQueue}, and \ref{lm:Q}, we conclude that $M(i)$ is the sum of $d_i$ and the minimum of the larger value between an increasing function $U$ and a decreasing queue $Q$ of $j$, for $i < j < n$ and $M(j) \in Q$.

To simplify the description, we use $j^*(i)$ to denote the $j$ that minimizes Equation~\ref{eq:TU}, as defined in Equation~\ref{eq:jstar}.
%To simplify the description, we use $j^*(i)$ in Equation~\ref{eq:jstar} to denote the $j$ that minimizes Equation~\ref{eq:T-rec-simple}.
This observation reduces the search for $j^*(i)$ in Equation~\ref{eq:TU} from considering each $U(i, j)$ and $M(j)$ for all $j$ between $i$ and $n$ to identifying the intersection of the increasing function $U$ and the decreasing queue $Q$ if they intersect.
%$j^*(i)$ will be at the intersection of the function $U$ and the values in $Q$ if they do intersect.
If the function values of $U$ and the values in $Q$ do not intersect, then the $j^*(i)$ will be $n$ or $i$.
Figure~\ref{fig:compute_M(2)} illustrates this observation.

\begin{equation}
j^*(i) = \arg \min_{i < j < n}(\max(M(j), U(i, j))) \label{eq:jstar}
\end{equation}

\subsubsection{The Second Observation}
%Next, we aim to establish a connection between the computation of $M(i)$ and that of $M(i - 1)$.
We observe how $U(i, j)$ changes when $i$ decreases for a fixed $j$.

\begin{eqnarray}
\lefteqn{U(i - 1, j)}  \label{eq:Ui1} \\
& = &  s(i - 1,j) + d_j \nonumber \\ 
& = & \sum_{k = i}^{j-1} d_k + \max_{i-1 \leq k < j}(d_k) + d_j \label{eq:U-update}\\
& = & (d_i + \sum_{k = i+1}^{j-1} d_k) + (\max(d_{i-1}, \max_{i \leq k < j}(d_k))) + d_j \nonumber\\
& > & \sum_{k = i+1}^{j-1} d_k + \max_{i \leq k < j}(d_k) + d_j \\
& = & s(i,j) + d_j \\
& = & U(i,j)
\end{eqnarray}

The inequality $U(i - 1, j) > U(i, j)$ indicates that after we compute $M(i)$ by finding $j^*(i)$ that minimizes Equation~\ref{eq:TU}, we only need to consider those $j$'s that are no greater than $j^*(i)$ when we compute $M(i - 1)$ through minimizing Equation~\ref{eq:TU}.
The dotted arrows in Figure~\ref{fig:compute_M(1)} illustrate the change from $U(2, j)$ to $U(1, j)$.
%The inequality $U(i -1, j) > U(i, j)$ indicates that after we compute $M(i)$ by finding $j^*(i)$ that minimizes Equation~\ref{eq:T-rec-simple}, we only need to consider those $j$'s that are no greater than $j^*(i)$ to minimize Equation~\ref{eq:T-rec-simple} for $M(i-1)$.

\input{tikz/linear_time_algorithm_Demo}

\begin{lemma} \label{lm:jstar}
$j^*$ is a increasing function of $i$, i.e., $j^*(i-1) \leq j^*(i)$.
\end{lemma}
\begin{proof}
In Equation~\ref{eq:TU}, $M(j)$ is a function of $j$ only and not a function of $i$.
However, for all $j$ from $i + 1$ to $n - 1$, $U(i - 1, j) > U(i, j)$, and $U(i - 1, j)$ is a increasing function of $j$.
As a result, if a $M(j) \in Q$ and $U(i-1, j)$ intersect, they will intersect at a $j$ value that is no greater than $j^*(i)$.
\end{proof}

Lemma~\ref{lm:jstar} leads to the second observation: to find the $j^*(i - 1)$, the index that minimizes $M(i - 1)$ in Equation~\ref{eq:TU}, we only need to consider those $j$'s that are no greater than $j^*(i)$.
%Lemma~\ref{lm:jstar} leads to the second observation: to find the $j$ that minimizes $M(i - 1)$, we only need to consider $j$'s that are no greater than $j^*(i)$.
In other words, when computing all $M(i)$ for $i$ from $n$ to $1$, we only need to find the $j$ that minimizes $M(i)$ in the direction of decreasing $j$.
%In other words, we will compute all $M(i)$ for $i$ from $n$ to $1$, and we only need to find the $j$ that minimizes $M(i)$ in the direction of decreasing $j$.

Figure~\ref{fig:computation_Demo} illustrates the second observation.
According to the first observation, the minimum is yielded at the intersection of the black stars and dots.
With the transition from $U(2,\cdot)$ to $U(1,\cdot)$, as indicated by the dotted arrows in Figure~\ref{fig:compute_M(1)}, we observe that the index of the intersection between the black stars and dots when computing $M(2)$ is no less than the index of their intersection when computing $M(1)$.
Therefore, we have $j^*(1) \leq j^*(2)$.

%The computation of Equation~\ref{eq:U-update} takes $O(1)$ time since we can compute $U(i-1, j)$ from $U(i, j)$ from Equation~\ref{eq:U-update} by tracking $\sum_{k = i+1}^{j-1} d_k$ and $\max_{i \leq k < j}(d_k)$, so that we can easily compute $d_i + \sum_{k = i + 1}^{j - 1} d_k$ and $\max(d_i, \max_{i + 1 \leq k \leq j - 1}(d_k))$ for a fixed $j$ in Equation~\ref{eq:U-update}.
% As a result, we can compute all $M(i)$'s with $O(n)$ time since we can compute every $M(i)$ in amortized $O(1)$ time.

\subsubsection{The Linear Time Algorithm}
We list the pseudo-code of the dynamic checkpoint selection problem in Algorithm~\ref{alg:dynamic-checkpoint-selection}.
%We list the pseudo-code of the dynamic checkpoint selection in Algorithm~\ref{alg:dynamic-checkpoint-selection}.

\begin{algorithm}[h!tb]
\begin{algorithmic}[1]
\caption{Dynamic Checkpoint Selection Problem in $O(n)$ time}
\label{alg:dynamic-checkpoint-selection}
\Require data sizes $d_1, \ldots, d_n$.
\Ensure The minimum cost of the dynamic checkpoint selection problem.
\State{$Q = \emptyset$}
\State{$j^{*} = n$}
\For{$i \gets n$ to $1$} \Comment{Compute $M(i)$ for $i$ from $n$ to $1$}
    \State{ $j = j^{*}$} \LineLabel{alg_line:set_j_from_jstar}
    \If{$i < n$}
        \State{Compute $U(i, j)$ from $U(i + 1, j)$ using Equation~\ref{eq:U-update}} \LineLabel{alg_line:Uij_to_Ui-1j}
    \EndIf
    \While {$U(i, j) \geq M(j)$ and $M(j)$ is not the first element in $Q$} \label{alg_line:while_start}\Comment{repeat until a crossover}
        \State {Set $j'$ to be the index of the previous element of $M(j)$ in $Q$}
        \State {Compute $U(i, j')$ from $U(i, j)$ using Equation~\ref{eq:U-(i,dot)}} \LineLabel{alg_line:Uij_to_Uij'}
        \State {$j = j'$} \LineLabel{alg_line:j_to_j'_inside_while}
%        \State {Compute $U(i, j)$ from $U(i + 1, j)$, as in Equation~\ref{eq:U-update}}
    \EndWhile \LineLabel{alg_line:while_end}
    \State {Set $M(i)$ to $d_i + \min (\max (M(j), U(i, j)), \max (M(j+1), U(i, j+1)))$}
    \State {Set $j^*$ to $j$ or $j + 1$ according to the minimum from the previous statement.}  \LineLabel{alg_line:set_j*}
    \State {Remove those $M$'s at the beginning of $Q$ that are no less than $M(i)$}
    \State {Add $M(i)$ at the beginning of $Q$}
\EndFor
\State{return $M(1)$} \Comment{the answer}
\end{algorithmic}
\end{algorithm}
\newpage
\begin{theorem}
The dynamic checkpoint selection problem is $O(n)$-time solvable, where $n$ is the number of neural network layers.
\begin{proof}
To prove the correctness, we show that during the $i$-th iteration, Algorithm~\ref{alg:dynamic-checkpoint-selection} obtains the correct $M(i)$.
According to Lemma~\ref{lm:jstar} in the second observation, the search for the $j^*(i)$ begins at $j^*(i + 1)$, which is implemented in line~\ref{alg_line:set_j_from_jstar}.
The first observation indicates that $j^*(i)$ occurs at the intersection of $U(i,\cdot)$ and $Q$.
The ``while'' loop from line~\ref{alg_line:while_start} to line~\ref{alg_line:while_end}, computing the first index $j$ before this intersection, implements this observation.
This is achieved since the function $U(i,\cdot)$ and the maintenance of $Q$ are the same as described in the first observation.
Subsequently, we assign $M(i)$ as the smaller value between $d_i + \max (M(j), U(i, j))$ and $d_i + \max (M(j + 1), U(i, j + 1))$ as indicated by Equation~\ref{eq:TU}.
%The reason for considering both $j$ and $j + 1$ is that the minimum value may occur at the first index after the intersection or the first index before the intersection.
The two subfigures of Figure~\ref{fig:computation_Demo} illustrate why both $j$, the first index before the intersection, and $j + 1$, the first index after the intersection, may be the index $j^*(i)$ that minimizes Equation~\ref{eq:TU}.
%In Figure~\ref{fig:compute_M(2)}, the index that minimizes $M(2)$ occurs at the first index after the intersection, while in Figure~\ref{fig:compute_M(1)}, the index that minimizes $M(1)$ occurs at the first index before the intersection.
We have thus proven the correctness of the algorithm~\ref{alg:dynamic-checkpoint-selection}.

We will now prove that the time complexity of Algorithm~\ref{alg:dynamic-checkpoint-selection} is $O(n)$.
The time complexity of Algorithm~\ref{alg:dynamic-checkpoint-selection} consists of the time to maintain $Q$ and to compute each $U(i, j)$ in line~\ref{alg_line:Uij_to_Ui-1j} and line~\ref{alg_line:Uij_to_Uij'}.

Each $M(i)$ will enter and exit $Q$ once, from either its head or its tail.
%From the construction of $Q$ we know that each $M(i)$ will enter and exit $Q$ once, from either its head or its tail.
Therefore, the time complexity of maintaining $Q$ is $O(n)$.

The computation of $U(i, j)$ at line~\ref{alg_line:Uij_to_Ui-1j} is performed $n - 1$ times.
Each $U(i, j)$ is derived from $U(i + 1, j)$ using Equation~\ref{eq:U-update}.
By maintaining $\displaystyle\sum_{k = i + 2}^{j - 1} d_k $ and $\displaystyle\max_{i + 1 \leq k < j}(d_k) $, the terms $\displaystyle\sum_{k = i + 1}^{j - 1} d_k $ and $\displaystyle\max_{i \leq k < j}(d_k) $ in Equation~\ref{eq:U-update} can be computed in $O(1)$ time.  
As a result, the computation of $U(i, j)$ at line~\ref{alg_line:Uij_to_Ui-1j} across all iterations takes O(n) time.

Each computation of $U(i, j')$ at line~\ref{alg_line:Uij_to_Uij'} takes $O(j - j')$ time by tracking $\displaystyle\sum_{k = i+1}^{j} d_k$ and $\displaystyle\max_{i \leq k \leq j}(d_k)$ in Equation~\ref{eq:U-(i,dot)}.
Moreover, after the computation, the value of $j$ is decreased by $j - j'$ (line~\ref{alg_line:j_to_j'_inside_while}).
The time complexity of the computation in line~\ref{alg_line:Uij_to_Uij'}, therefore, depends on the range by which $j$ is reduced.
Initially, $j$ equals $n$.
It remains positive throughout the computation, and in each iteration of the ``for'' loop, the value of $j$ can increase by at most $1$ (line~\ref{alg_line:set_j*}).
Therefore, the range by which $j$ is reduced is bounded by $2n$, implying the $O(n)$ time complexity of the computation in line~\ref{alg_line:Uij_to_Uij'}.
%From Lemma~\ref{lm:jstar} we only need to consider $j$ in decreasing order when we compute $M(i)$ in Equation~\ref{eq:TU} in decreasing $i$ order.
%There are only $n$ possible $j$ values, and each update in Equation~\ref{eq:U-update} takes $O(1)$ time.

As a result, the time complexity of Algorithm~\ref{alg:dynamic-checkpoint-selection} is $O(n)$.
\end{proof}
\end{theorem}
%}

%% file: tikz/linear_time_algorithm_Demo.tex
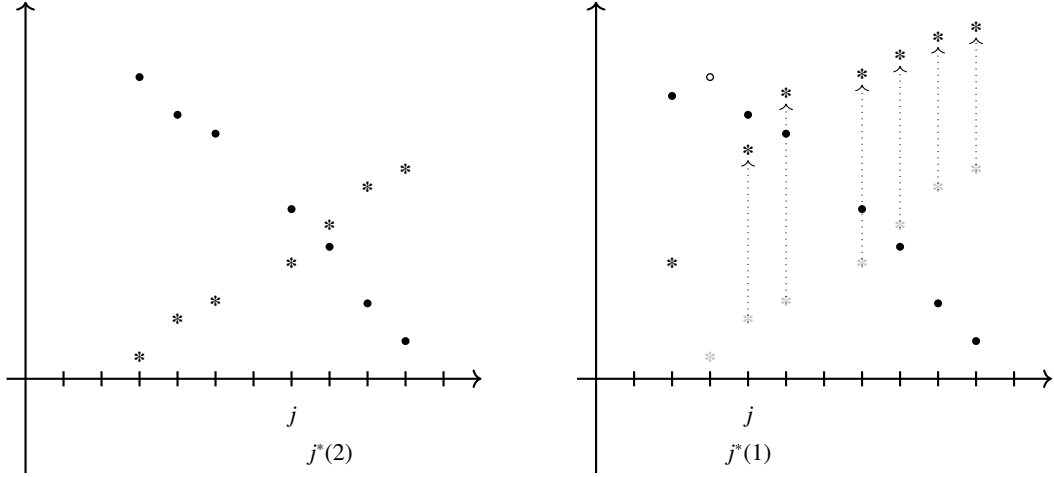
\begin{figure}
    \centering
    \begin{subfigure}[t]{.45\textwidth}
        \input{tikz/linear_time_iteration_i}
        \caption{The values of $M(\cdot)$ and $U(2, \cdot)$ when we compute $M(2)$.}
        \label{fig:compute_M(2)}
    \end{subfigure}
    \begin{subfigure}[t]{.45\textwidth}
        \input{tikz/linear_time_iteration_i-1} 
        \caption{The values of $M(\cdot)$ and $U(1, \cdot)$ when we compute $M(1)$.
        The gray stars are the values of $U(2,\cdot)$.
        The hollow dot represents the element at the head of $Q$ larger than $M(2)$ and is therefore removed from $Q$ before $M(2)$ is inserted.}
        \label{fig:compute_M(1)}
    \end{subfigure}
    \caption{The values of $M(\cdot)$ and $U(i, \cdot)$ when we compute $M(2)$ and $M(1)$.
    The $x$-coordinate represents the index of $M(\cdot)$ and $U(i, \cdot)$, while the $y$-coordinate represents their corresponding values.
    Black dots indicate the values of $M(\cdot)$ present in the queue $Q_i$, and stars represent the values of $U(i, \cdot)$.
    The $x$-coordinate $i$ without a black dot indicates that $M(i)$ has been removed from the queue $Q$.
    The $j$ below the $x$-axis marks the value of $j$ after the ``while'' loop in Algorithm 6.
    The $j^*(i)$ below the $x$-axis marks the value of updated $j^*_(i)$.
    }
\label{fig:computation_Demo}
\end{figure}

%% file: tikz/linear_time_iteration_i.tex
\begin{tikzpicture}[scale=0.5]
    \draw[->, thick] (-0.5, 0) -- (12, 0);
    \draw[->, thick] (0, -2.5) -- (0, 10);

    \draw[thick] (1, -0.2) -- (1, 0.2);
    \draw[thick] (2, -0.2) -- (2, 0.2);
    \draw[thick] (3, -0.2) -- (3, 0.2);
    \draw[thick] (4, -0.2) -- (4, 0.2);
    \draw[thick] (5, -0.2) -- (5, 0.2);
    \draw[thick] (6, -0.2) -- (6, 0.2);
    \draw[thick] (7, -0.2) -- (7, 0.2);
    \draw[thick] (8, -0.2) -- (8, 0.2);
    \draw[thick] (9, -0.2) -- (9, 0.2);
    \draw[thick] (10, -0.2) -- (10, 0.2);
    \draw[thick] (11, -0.2) -- (11, 0.2);

    % Dots for M values
    \fill[black] (10, 1) circle (3pt);
    \fill[black] (9, 2) circle (3pt);
    \fill[black] (8, 3.5) circle (3pt);
    \fill[black] (7, 4.5) circle (3pt);
    \fill[black] (5, 6.5) circle (3pt);
    \fill[black] (4, 7) circle (3pt);
    \fill[black] (3, 8) circle (3pt);

    % Stars for U values
    \node[scale=0.75] at (3, 0.5) {\textbf{*}};
    \node[scale=0.75] at (4, 1.5) {\textbf{*}};
    \node[scale=0.75] at (5, 2) {\textbf{*}};
    \node[scale=0.75] at (7, 3) {\textbf{*}};
    \node[scale=0.75] at (8, 4) {\textbf{*}};
    \node[scale=0.75] at (9, 5) {\textbf{*}};
    \node[scale=0.75] at (10, 5.5) {\textbf{*}};

    \node[scale=0.75] at (7, -1) {$j$};
    \node[scale=0.75] at (8, -2) {$j^*(2)$};
\end{tikzpicture}
\comment{The values of $M(\cdot)$ and $U(i, \cdot)$ at the $i$-th iteration.
The $x$-coordinate represents the index of $M(\cdot)$ and $U(i, \cdot)$, while the $y$-coordinate represents their corresponding values.
Black dots indicate the values of $M(\cdot)$ present in the queue $Q_i$, and stars represent the values of $U(i, \cdot)$.
The $j$ below the $x$-axis marks the value of $j$ after the ``while'' loop in Algorithm~?.
The $j^*$ below the $x$-axis marks the value of updated $j^*$ at the $i$-th iteration.
}

%% file: tikz/linear_time_iteration_i-1.tex
\begin{tikzpicture}[scale=0.5]
    \draw[->, thick] (-0.5, 0) -- (12, 0);
    \draw[->, thick] (0, -2.5) -- (0, 10);

    \draw[thick] (1, -0.2) -- (1, 0.2);
    \draw[thick] (2, -0.2) -- (2, 0.2);
    \draw[thick] (3, -0.2) -- (3, 0.2);
    \draw[thick] (4, -0.2) -- (4, 0.2);
    \draw[thick] (5, -0.2) -- (5, 0.2);
    \draw[thick] (6, -0.2) -- (6, 0.2);
    \draw[thick] (7, -0.2) -- (7, 0.2);
    \draw[thick] (8, -0.2) -- (8, 0.2);
    \draw[thick] (9, -0.2) -- (9, 0.2);
    \draw[thick] (10, -0.2) -- (10, 0.2);
    \draw[thick] (11, -0.2) -- (11, 0.2);

    % Stars for previous U values
    \node[scale=0.75, color=lightgray] at (3, 0.5) {\textbf{*}};
    \node[scale=0.75, color=lightgray] at (4, 1.5) {\textbf{*}};
    \node[scale=0.75, color=lightgray] at (5, 2) {\textbf{*}};
    \node[scale=0.75, color=lightgray] at (7, 3) {\textbf{*}};
    \node[scale=0.75, color=lightgray] at (8, 4) {\textbf{*}};
    \node[scale=0.75, color=lightgray] at (9, 5) {\textbf{*}};
    \node[scale=0.75, color=lightgray] at (10, 5.5) {\textbf{*}};

    % Dots for M values
    \fill[black] (10, 1) circle (3pt);
    \fill[black] (9, 2) circle (3pt);
    \fill[black] (8, 3.5) circle (3pt);
    \fill[black] (7, 4.5) circle (3pt);
    \fill[black] (5, 6.5) circle (3pt);
    \fill[black] (4, 7) circle (3pt);
    \fill[black] (3, 8) circle (3pt);
    \fill[white] (3, 8) circle (2pt);
    \fill[black] (2, 7.5) circle (3pt);

    % Stars for U values
    \node[scale=0.75] at (2, 3) {\textbf{*}};
    \node[scale=0.75] at (4, 6) {\textbf{*}};
    \node[scale=0.75] at (5, 7.5) {\textbf{*}};
    \node[scale=0.75] at (7, 8) {\textbf{*}};
    \node[scale=0.75] at (8, 8.5) {\textbf{*}};
    \node[scale=0.75] at (9, 9) {\textbf{*}};
    \node[scale=0.75] at (10, 9.25) {\textbf{*}};

    \draw[-{>[scale=1]}, dotted] (4, 1.5) -- (4, 6-0.2);
    \draw[-{>[scale=1]}, dotted] (5, 2) -- (5, 7.5-0.2);
    \draw[-{>[scale=1]}, dotted] (7, 3) -- (7, 8-0.2);
    \draw[-{>[scale=1]}, dotted] (8, 4) -- (8, 8.5-0.2);
    \draw[-{>[scale=1]}, dotted] (9, 5) -- (9, 9-0.2);
    \draw[-{>[scale=1]}, dotted] (10, 5.5) -- (10, 9.25-0.2);

%    \node[scale=1] at (8, -1) {$j$};
%    \node[scale=1] at (4, -2) {$j$};
%    \node[scale=1] at (4, -3) {$j*$};
    \node[scale=0.75] at (4, -1) {$j$};
    \node[scale=0.75] at (4, -2) {$j^*(1)$};
\end{tikzpicture}
\comment{The distribution of $M(\cdot)$ and $U(i, \cdot)$ during the $(i - 1)$-th iteration.
The head of $Q$ is removed from the queue (the hollow black dot) since $M(i)$ (the black dot on the left of the hollow black dot) is smaller than it.
All the $U(i, \cdot)$ (the light gray stars) become the $U(i - 1, \cdot)$ (the black stars).
Note that we don't have to compute every values of $U(i,\cdot)$.
The first $j$ below the $x$-axis marks the value of $j$ before the ``while'' loop at the $(i - 1)$-th iteration.
In other words, it equals the $j^*$ at the $i$-th iteration.
The second $j$ below the $x$-axis marks the value of $j$ after the ``while'' loop at the $(i - 1)$-th iteration.
The $j^*$ below the $x$-axis marks the value of $j^*$ at the $(i - 1)$-th iteration.
}

%% file: 5_experiment.tex
\section{Experiment} \label{sec:experiment}

In this section, we evaluate our algorithms presented in the previous sections. We profile the GPU memory usage when training two linear feedforward DNN models with the optimal checkpoint subset returned by the algorithms. We do the same for comparison by profiling the training of the same models with non-optimal checkpoint subsets. Finally, we run the algorithm implementation of Feng et al.~\cite{feng2021optimal} to find and compare the differences between their works and ours.

\subsection{Environment Settings}

\subsubsection{Models and Definitions}

We use VGG-19~\cite{Simonyan2014VeryDC} and AlexNet~\cite{krizhevsky2017imagenet} as our DNN model to evaluate our algorithm.
VGG-19 has 16 convolution layers and 5 max-pooling layers, and 3 fully connected layers.
AlexNet has 5 convolution layers and 3 max-pooling layers, and 3 fully connected layers.

\begin{figure}[h!]
    \centering
    \includegraphics[width=\textwidth]{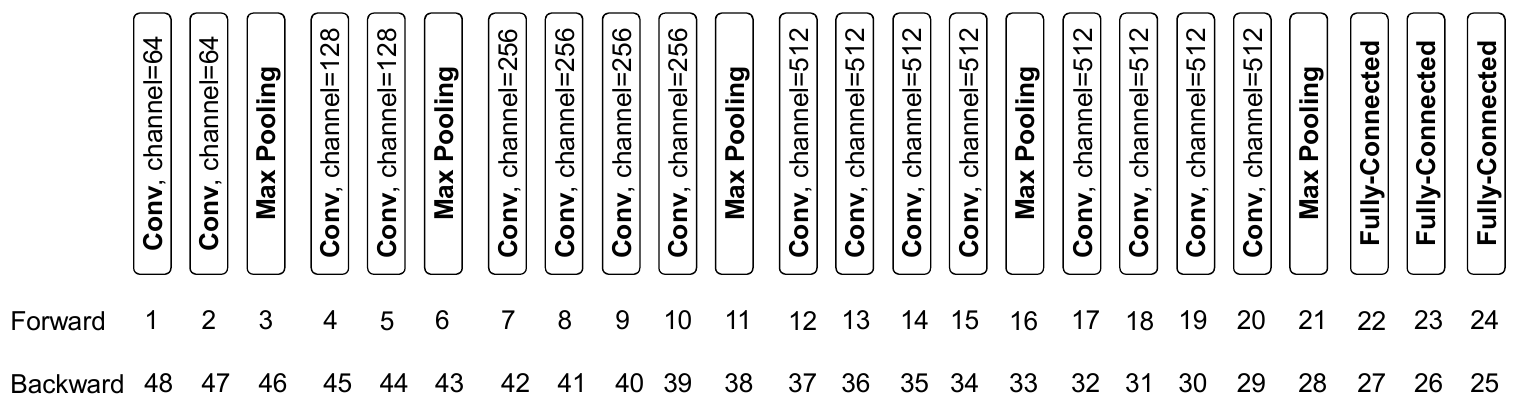}
    \caption{Training phase indices of VGG-19.} 
    \label{fig:vgg19_layers}
\end{figure}

\begin{figure}[h!]
    \centering
    \includegraphics[width=.7\textwidth]{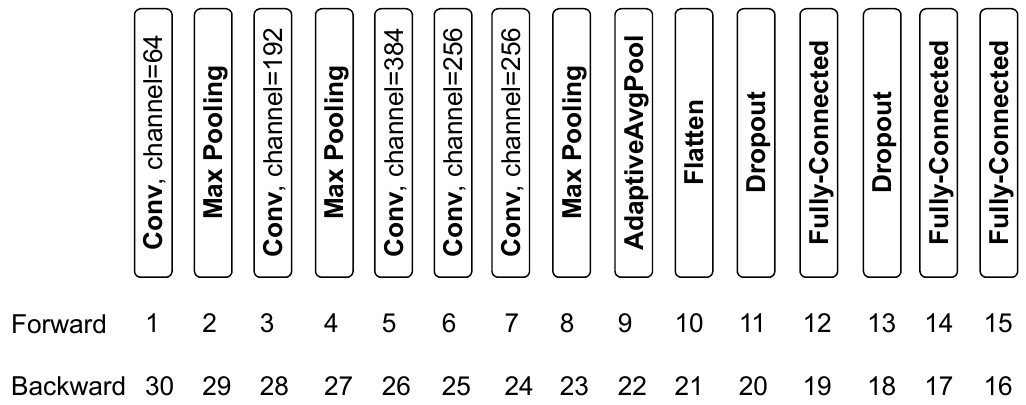}
    \caption{Training phase indices of AlexNet.} 
    \label{fig:alexnet_layers}
\end{figure}

We define the term {\em training phase index} to describe every stage of training, including both the forward and backward phases. It is a number ranging from 0 to $2N+1$, where $N$ is the number of layers of the model. The value $0$ represents the stage before the training, and $2N+1$ represents the stage after training.
The training is in the forward phase when $i$ is in the range $1 \leq i \leq N$, and it is in the backward phase when $i$ is in the range $N+1 \leq i \leq 2N$. Now we can index the entire training.
Notice that we always measure the GPU memory at the end of every stage. For ease of index calculation, we do not merge the forward phase and the backward phase of the last layer.
Figures~\ref{fig:vgg19_layers} and \ref{fig:alexnet_layers} depict the training phase indices for VGG-19 and AlexNet, respectively. For example, the training phase index 29 of VGG-19 means that it is the stage in the backward phase of the 20th layer, i.e. we are in the backward phase of the last convolutional layer with 512 output channels.


\subsubsection{Benchmarks}

%We use ImageNet~\cite{russakovsky2015imagenet} and TinyImageNet to evaluate our algorithm.
%ImageNet consists of around 1.3 million color images of different sizes in 1000 classes.
%TinyImageNet~ consists of 0.1 million color images of the same size (64 by 64) in 200 classes.

%For ImageNet, we will crop all images into size 224 by 224 before feeding them into our model.

We use ImageNet~\cite{russakovsky2015imagenet} to evaluate our algorithms.
ImageNet consists of around 1.3 million color images of different sizes in 1000 classes.
We crop all images into size 224 by 224 before feeding them into the model.

\subsubsection{Implementation}

We use PyTorch~\cite{paszke2019pytorch} deep learning platform to implement the two DNN models VGG-19 and AlexNet.
To implement the recomputing of a segment in the checkpoint technique, we use the PyTorch API \texttt{torch.utils.checkpoint}.
To profile the GPU memory usage, we use the PyTorch API \texttt{torch.cuda.memory\_allocated}.
We include the fully-connected layers of both VGG-19 and AlexNet when selecting the candidate of checkpoints, in addition to convolution layers. 
This is because the memory cost of fully-connected layers is large.
Additionally, we exclude the layers of AdaptiveAvgPool2d, Flatten, and Dropout in VGG-19, but include these layers as checkpoint candidates in AlexNet.
This simplifies the selection of checkpoints for the $O(\sqrt{n})$ algorithm~\cite{chen2016training}.
We perform all experiments with GPUs of the same specs, NVIDIA GeForce RTX 3090, 24 GiB memory on a server with 16-core, 2.90 GHz Intel Xeon Gold 6226R CPU, 192 GiB RAM.

\subsection{Algorithm Prediction versus PyTorch Report} \label{sec:prediction}

In this experiment, we demonstrate the importance of the consideration of the output gradient buffer as we have mentioned in section~\ref{sec:dynamic_checkpoint}. We show that the prediction of GPU memory usage by our algorithm is aligned with the report from PyTorch.

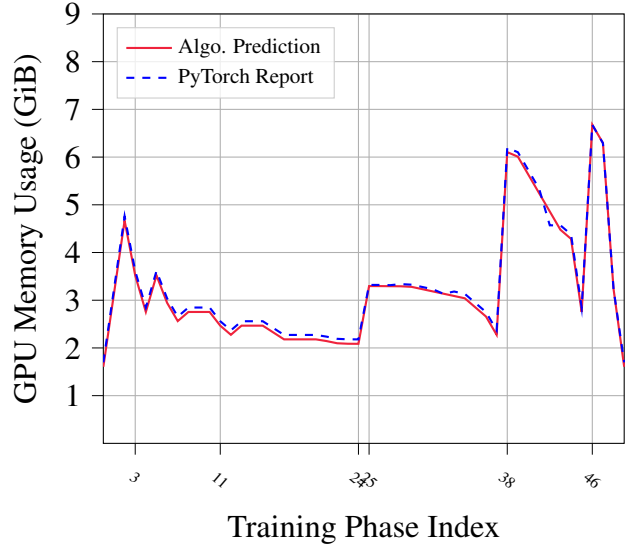
\begin{figure}[h!t]
    \centering
    \input{tikz/algo_pred_vs_pytorch_report}
    \caption{GPU Memory Usage: Algorithm Prediction vs PyTorch Report on VGG-19 with Checkpoint Subset \{3, 11, 24\}.} 
    \label{fig:algo_pred_vs_pytorch_report}
\end{figure}

In Figure~\ref{fig:algo_pred_vs_pytorch_report}, we profile our training of VGG-19 with the checkpoint subset returned by our algorithm and draw the GPU memory usage prediction using the same checkpoint subset.
The blue dashed line represents the GPU memory usage reported by PyTorch.
On the other hand, the red line is the GPU memory usage predicted by our algorithm. Notice that we use the checkpoint subset \{3, 11, 24\} reported by our algorithm when running on VGG-19 to mark the corresponding stages on the x-axis for both the forward and backward phases, so the six ticks on the x-axis are all stages about checkpoints.
The average error of our prediction is 2.8\%.

\subsection{Comparison with $O(\sqrt{n})$ Memory Cost Algorithm}

In this experiment, we compare the GPU memory usage when training VGG-19 using the checkpoint subset provided by our algorithm and the one by the sublinear memory cost algorithm proposed by Chen et al.

\begin{figure}[h!tb]
    \centering
    \input{tikz/our_algo_vs_sublinear}
    \caption{GPU Memory Usage: Our Algorithm vs $O(\sqrt{n})$ Memory cost Algorithm Proposed by Chen et al. on VGG-19} 
    \label{fig:our_algo_vs_sublinear}
\end{figure}
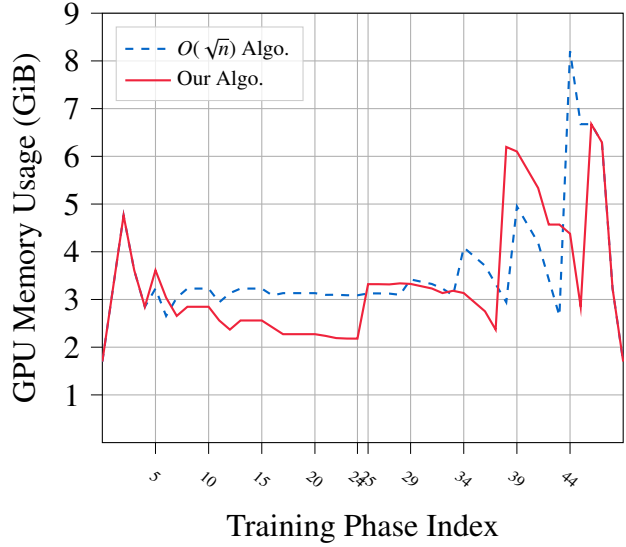

In Figure~\ref{fig:our_algo_vs_sublinear}, the red line represents the GPU memory usage of the training on VGG-19 reported by PyTorch using the checkpoint subset returned by our algorithm. On the other hand, the blue dashed line represents the GPU memory usage of the training using the checkpoint subset calculated by the sublinear memory cost algorithm. We mark the ticks of the x-axis using the checkpoint subset of the latter, which is the set \{5, 10, 15, 20, 24\}.

\subsection{Comparison with Optimal Arbitrary Computation Graph Algorithm}

In this experiment, we compare the GPU memory usage when training VGG-19 using the checkpoint subset provided by our algorithm and the one by the optimal arbitrary computation graph algorithm proposed by Feng et al.

\begin{figure}[h!tb]
    \centering
    \input{tikz/our_algo_vs_optimal_ACG}
    \caption{GPU Memory Usage: Our Algorithm vs Optimal Arbitrary Computation Graph (ACG) Algorithm Proposed by Feng et al. on VGG-19} 
    \label{fig:our_algo_vs_optimal_ACG}
\end{figure}
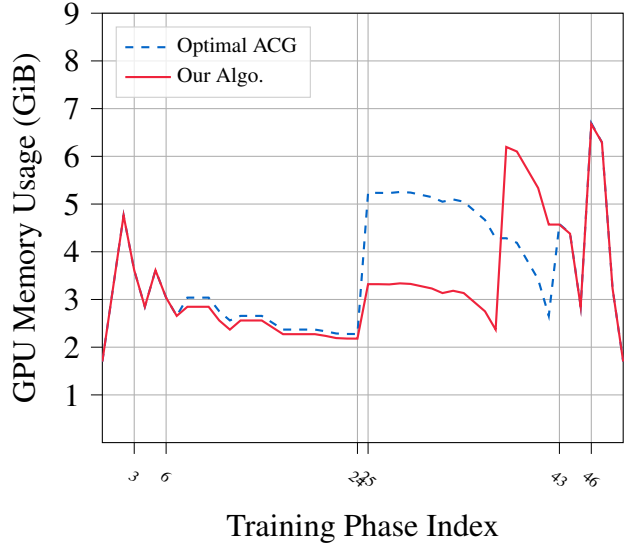

In Figure~\ref{fig:our_algo_vs_optimal_ACG}, the red line represents the GPU memory usage of the training on VGG-19 reported by PyTorch using the checkpoint subset returned by our algorithm. On the other hand, the blue dashed line represents the GPU memory usage of the training using the checkpoint subset returned by the optimal arbitrary computation graph algorithm. We mark the ticks of the x-axis using the checkpoint subset of the latter, which is the set \{3, 6, 24\}.
In the figure, we can see that both algorithms have the same peak memory usage that occurs at the start of the last segment in the backward phase. During the same training phase index range from 25 to 43, the training using the optimal ACG algorithm spends around 2 GiB more memory compared to our algorithm on the first half and more of the first segment during the backward phase. On the other hand, our algorithm also spends around 2 GiB more memory compared to the optimal ACG algorithm on the remaining part of the first segment during the backward phase.

\subsection{Comparison: The Peak Memory Usage on Different Experiment Settings}

\begin{table*}[h!tb]
\centering
\caption{The Profiling Results of Peak Memory Usage: Finding the Checkpoint Subset Using Different Methods}
\label{tab:PR_Peak_Memory_Usage}
\footnotesize
\begin{tabular}{|c|c|c|c|c|c|}
    \hline
    Peak Mem.(MiB) & Algorithm~\ref{alg:dynamic-checkpoint-selection} & Algorithm~\ref{alg:checkpoint-selection} & ACG Solver & $O(\sqrt{n})$ & PyTorch \\
    \hline
    \hline
%    VGG-19, b=128 & {6,444 \{2,4,6,9,11,14,\par 16,19,21,23,24\}} & 6,835 \{3,6,24\} & 6,836 \{3,6,24\} & {8,404 \{5,10,15,\par 20,24\}} & 11,262 \\
    VGG-19, b=128 & 6,444 & 6,835 & 6,836 & 8,404 & 11,262 \\
    & {\{2,4,6,9,11,14,\par 16,19,21,23,24\}} & \{3,6,24\} & \{3,6,24\} & {\{5,10,15,\par 20,24\}} & \\
    \hline
    VGG-19, b=256 & 11,220 & 12,004 & 12,004 & 15,139 & 20,850 \\
    \hline
    VGG-19, b=320 & 13,609 & 14,592 & 14,592 & 18,511 & OOM \\
    \hline
    VGG-19, b=384 & 15,997 & 17,177 & 17,177 & OOM & OOM \\
    \hline
    VGG-19, b=400 & 16,594 & 17,824 & 17,824 & OOM & OOM \\
    \hline
    \hline
%    AlexNet, b=128 & 1,002 \{2,4,6,8\} & 1,003 \{2,4,12,15\} & 1,003 \{2,4,12,15\} & 1,106 \{4,8,12,15\} & 1,174 \\
    AlexNet, b=128 & 1,002 & 1,003 & 1,003 & 1,106 & 1,174 \\
    & \{2,4,6,8,12,14,15\} & \{2,4,12,15\} & \{2,4,12,15\} & \{4,8,12,15\} & \\
    \hline
    AlexNet, b=4096 & 9,866 & 9,866 & 9,866 & 13,182 & 15,275 \\
    \hline
\end{tabular}
\end{table*}

In this experiment, we compare the GPU memory usage when training both VGG-19 and AlexNet with different batch sizes and checkpoint subsets that are computed using different algorithms. The results are shown in Table~\ref{tab:PR_Peak_Memory_Usage}.
In the first column, the shorthand b=$N$ means that we use the number $N$ as the batch size for all experiments of the current row. The abbreviation OOM means that the experiment cannot be carried out due to the out-of-memory error reported by PyTorch.
For the benchmark, we use a single batch from the ImageNet database for testing. (In the implementation of Feng et al., they randomly generate a batch of images for testing.)

For the VGG-19 results in Table~\ref{tab:PR_Peak_Memory_Usage}, both algorithm~\ref{alg:checkpoint-selection} and the optimal ACG algorithm have the same peak memory usage reported by PyTorch since the checkpoint subsets returned by the two algorithms are the same. On the other hand, algorithm~\ref{alg:dynamic-checkpoint-selection} has the lowest peak GPU memory usage. The checkpoint subset returned by algorithm~\ref{alg:checkpoint-selection} and the optimal ACG algorithm is \{3, 6, 24\}. The checkpoint subset returned by algorithm~\ref{alg:dynamic-checkpoint-selection} is \{2, 4, 6, 9, 11, 14, 16, 19, 21, 23, 24\}.
Compared to the $O(\sqrt{n})$ algorithm and ACG Solver, algorithm~\ref{alg:dynamic-checkpoint-selection} reduces 2 GiB (23\%) and 392 MiB (5.7\%) of peak memory usage with a batch size of 128, respectively.
More importantly, the checkpoint subsets do not change when we increase the batch size since the batch size is a factor in the output data size of every layer.

\subsection{The Granularity of Model Specs Matters: Testing The Optimal Checkpoints on AlexNet with Extended Model Specs}

In this subsection, we run Algorithm~\ref{alg:dynamic-checkpoint-selection} on AlexNet model to get the optimal checkpoint subset, and train and profile AlexNet with these checkpoints accordingly. From our prior experiments with VGG-19 models, we observe that the peak GPU memory usage usually occurs at one of the max-pooling layers. But in the plain model specs of AlexNet, all of its max-pooling layers are embedded into {\em larger, abstract} convolution layers, i.e. it is {\em hidden} from the selection of checkpoints. We move all of these max-pooling layers out of the convolution layers since the behavior of our algorithm is dependent on the granularity, i.e. the length, of the model specs. We expect that the peak GPU memory usage will decrease with the new model specs because more combinations are considered.

\subsubsection{AlexNet: Using the Plain Specs}

In this trial, we use the plain model specs of AlexNet, i.e. it has 5 convolution layers and 3 fully-connected layers. The result in Figure~\ref{fig:GMU_AlexNet_Ckpt_Optimal} indicates that our algorithm chose checkpoints \{3, 4, 5, 9, 11, 12\}. This means that the third and fourth convolution layers (Conv2d/ReLU), the fifth convolution layer (Conv2d/ReLU + MaxPool2d), and
%fourth(Conv2d(in=384, out=256), relu, MaxPool2d)  convolution layers
all three fully-connected layers are chosen as checkpoints.

\input{tikz/alexnet_optimal}

\subsubsection{AlexNet: Standalone Max-Pooling Layers}

\input{tikz/alexnet_optimal2}

In this trial, we use the extended model specs of AlexNet, i.e. it has 5 convolution layers and 3 standalone max-pooling layers, and 3 fully-connected layers. From the result in Figure~\ref{fig:GMU_AlexNet_Ckpt_Optimal2}, we can see that the checkpoint subset \{2, 4, 6, 8, 12, 14, 15\} is chosen by our algorithm.
%At first look, it might seem that the first checkpoint did not change, but
We should note that since we have changed the granularity of the specs, the numbers (which are indexes) now have a different meaning. It means that the three standalone max-pooling layers (i.e. indices 2, 4, and 8) are chosen as the checkpoints.

In the first convolution layer (Conv2d/ReLU + MaxPool2d) of AlexNet, the output tensor sizes of Conv2d/ReLU and MaxPool2d are 103 MiB and 26 MiB for a batch size of 128, respectively.
Without separating the max-pooling, only the tensor of size 26 MiB can be selected as the checkpoint.
By moving max-pooling out of the layer, the tensor of size 103 MiB becomes a selectable checkpoint.
Consequently, our algorithm identifies a better solution by selecting it as a checkpoint, reducing the peak memory usage by 100 MiB compared to the plain AlexNet model.
The result aligns with our expectations.
We can have a lower peak memory usage if there are more candidates in the model specs as if checkpointing at the middle of an abstract, high-level layer. 

This also shows the advantage of our algorithm. Because the time complexity of our algorithm is linear, the increase in model length caused by the increase in granularity will not affect the speed of our algorithm a lot. (in our cases, there will be at most 7 max-pooling layers given the image size is 224 by 224)

\subsection{Comparison: The Training Time on Different Experiment Settings}

\begin{table*}[h!tb]
\centering
\caption{The Training Time of One Batch Using Different Methods}
\label{tab:training_time}
\footnotesize
\begin{tabular}{|c|c|c|c|c|c|}
    \hline
    Exec. time per batch & Algorithm~\ref{alg:dynamic-checkpoint-selection} & Algorithm~\ref{alg:checkpoint-selection} & ACG Solver & $O(\sqrt{n})$ & PyTorch \\
     (second) & & & & & \\
    \hline
    \hline
%    VGG-19, b=128 & {6,444 \{2,4,6,9,11,14,\par 16,19,21,23,24\}} & 6,835 \{3,6,24\} & 6,836 \{3,6,24\} & {8,404 \{5,10,15,\par 20,24\}} & 11,262 \\
    VGG-19, b=128 & 0.779 & 0.779 & 0.779 & 0.780 & 0.585 \\
    & {\{2,4,6,9,11,14,\par 16,19,21,23,24\}} & \{3,6,24\} & \{3,6,24\} & {\{5,10,15,\par 20,24\}} & \\
    \hline
    VGG-19, b=256 & 1.541 & 1.540 & 1.540 & 1.541 & 1.158 \\
    \hline
    VGG-19, b=320 & 1.921 & 1.920 & 1.920 & 1.922 & OOM \\
    \hline
    VGG-19, b=384 & 2.292 & 2.289 & 2.289 & OOM & OOM \\
    \hline
    VGG-19, b=400 & 2.849 & 2.335 & 2.335 & OOM & OOM \\
    \hline
    \hline
%    AlexNet, b=128 & 1,002 \{2,4,6,8\} & 1,003 \{2,4,12,15\} & 1,003 \{2,4,12,15\} & 1,106 \{4,8,12,15\} & 1,174 \\
    AlexNet, b=128 & 0.046 & 0.046 & 0.046 & 0.046 & 0.038 \\
    & \{2,4,6,8,12,14,15\} & \{2,4,12,15\} & \{2,4,12,15\} & \{4,8,12,15\} & \\
    \hline
    AlexNet, b=4096 & 1.206 & 1.206 & 1.206 & 1.216 & 1.058 \\
    \hline
\end{tabular}
\end{table*}

In this experiment, we compare the training time of VGG-19 and AlexNet with different algorithms.
We train the model for one epoch on the ImageNet dataset and measure the average training time of a single batch.
The results are shown in Table~\ref{tab:training_time}.

For VGG-19, the training time is close with all checkpoint selection methods, irrespective of the number of segments.
The only exception is algorithm~\ref{alg:dynamic-checkpoint-selection} with a batch size of 400, which exhibits 22\% longer time compared to other checkpoint selection methods.
Since this paper focuses on finding the optimal checkpoints with minimal peak memory usage, we will investigate its reason in the future.
Compared with PyTorch, the checkpoint selection methods incur 33\% additional training time with batch sizes of 128 and 256.

As for AlexNet, since AlexNet is a relatively small model, data augmentation becomes a bottleneck and the GPU often becomes idle when waiting for the input data from the CPU.
Therefore, we disable data augmentation and reuse the same input batch data when measuring the training time of AlexNet.
This way avoids extra overhead and ensures full GPU utilization.
The result in Table~\ref{tab:training_time} indicates that the checkpointing methods require 14\% more training time compared to PyTorch with a batch size of 4096.

%For AlexNet, the training time is close with all methods.
%This is because data augmentation is enabled for training, where the augmentation process runs on the CPU.
%Since AlexNet is a relatively small model, this data augmentation becomes a bottleneck, and it often causes the GPU to be idle when waiting for the input data.
%We conduct another experiment without enabling data augmentation and reuse the same input batch data for training AlexNet. This way avoids extra overhead and ensures full GPU utilization. The result indicates that all checkpointing methods require 14\% more training time compared to PyTorch without checkpointing.

We implemented our checkpoint selection algorithms using Python. Algorithm~\ref{alg:checkpoint-selection} and algorithm~\ref{alg:dynamic-checkpoint-selection} respectively take 20 milliseconds and 1.1 milliseconds to find the checkpoint subset for VGG-19, executed on the Intel Xeon Gold 6226R CPU. 

%% file: tikz/algo_pred_vs_pytorch_report.tex
% This file was created with tikzplotlib v0.10.1.
\begin{tikzpicture}

\definecolor{crimson2393560}{RGB}{239,35,60}
\definecolor{darkgray176}{RGB}{176,176,176}
\definecolor{lightgray204}{RGB}{204,204,204}

\begin{axis}[
legend cell align={left},
legend style={fill opacity=0.8, draw opacity=1, text opacity=1, draw=lightgray204},
tick align=outside,
tick pos=left,
title={ },
x grid style={darkgray176},
xlabel={Training Phase Index},
xmajorgrids,
xmin=0, xmax=49,
xtick style={color=black},
xticklabel style={rotate=340.0},
y grid style={darkgray176},
ytick={1,2,3,4,5,6,7,8,9},
xtick={3,11,24,46,38,25},
xticklabel style={rotate=340.0, font=\tiny},
legend style={font=\scriptsize},
legend style={at={(0,1)},anchor=north west, xshift=5pt,yshift=-5pt},
ylabel={GPU Memory Usage (GiB)},
ymajorgrids,
ymin=0, ymax=9,
ytick style={color=black}
]
\addplot [thick, crimson2393560]
table {%
0 1.605439453125
1 3.136689453125
2 4.667939453125
3 3.519501953125
4 2.753876953125
5 3.519501953125
6 2.945283203125
7 2.562470703125
8 2.753876953125
9 2.753876953125
10 2.753876953125
11 2.466767578125
12 2.275361328125
13 2.466767578125
14 2.466767578125
15 2.466767578125
16 2.323212890625
17 2.179658203125
18 2.179658203125
19 2.179658203125
20 2.179658203125
21 2.14376953125
22 2.09787109375
23 2.087861328125
24 2.08638671875
25 3.2965901184082
26 3.29611328125
27 3.29416015625
28 3.29220703125
29 3.280244140625
30 3.232392578125
31 3.184541015625
32 3.136689453125
33 3.088837890625
34 3.040986328125
35 2.849580078125
36 2.658173828125
37 2.275361328125
38 6.1039631652832
39 6.0082600402832
40 5.6254475402832
41 5.2426350402832
42 4.8598225402832
43 4.4770100402832
44 4.2856037902832
45 2.7543537902832
46 6.6781819152832
47 6.2953694152832
48 3.2328694152832
49 1.605439453125
};
\addlegendentry{Algo. Prediction}
\addplot [thick, blue, dashed]
table {%
0 1.69838619232178
1 3.22963619232178
2 4.76088619232178
3 3.61244869232178
4 2.84682369232178
5 3.61244869232178
6 3.03822994232178
7 2.65541744232178
8 2.84682369232178
9 2.84682369232178
10 2.84682369232178
11 2.55971431732178
12 2.36830806732178
13 2.55971431732178
14 2.55971431732178
15 2.55971431732178
16 2.41615962982178
17 2.27260494232178
18 2.27260494232178
19 2.27260494232178
20 2.27260494232178
21 2.23671627044678
22 2.19081783294678
23 2.18080806732178
24 2.17933177947998
25 3.3209228515625
26 3.3204460144043
27 3.3158073425293
28 3.3377799987793
29 3.3258171081543
30 3.2779655456543
31 3.2301139831543
32 3.1344108581543
33 3.1822624206543
34 3.1344108581543
35 2.9430046081543
36 2.7515983581543
37 2.3687858581543
38 6.1969108581543
39 6.1012077331543
40 5.7183952331543
41 5.3355827331543
42 4.5699577331543
43 4.5699577331543
44 4.3785514831543
45 2.8473014831543
46 6.6754264831543
47 6.2926139831543
48 3.2301139831543
49 1.69886350631714
};
\addlegendentry{PyTorch Report}
\end{axis}

\end{tikzpicture}

%% file: tikz/our_algo_vs_sublinear.tex
% This file was created with tikzplotlib v0.10.1.
\begin{tikzpicture}

\definecolor{crimson2393560}{RGB}{239,35,60}
\definecolor{darkgray176}{RGB}{176,176,176}
\definecolor{lightgray204}{RGB}{204,204,204}
\definecolor{royalblue4102200}{RGB}{4,102,200}

\begin{axis}[
legend cell align={left},
legend style={fill opacity=0.8, draw opacity=1, text opacity=1, draw=lightgray204},
tick align=outside,
tick pos=left,
title={ },
x grid style={darkgray176},
xlabel={Training Phase Index},
xmajorgrids,
xmin=0, xmax=49,
xtick style={color=black},
xticklabel style={rotate=340.0},
ytick={1,2,3,4,5,6,7,8,9},
xtick={5,10,15,20,24,44,39,34,29,25}, % 3,11,24,46,38,25,
xticklabel style={rotate=340.0, font=\tiny},
legend style={font=\scriptsize},
legend style={at={(0,1)},anchor=north west, xshift=5pt,yshift=-5pt},
y grid style={darkgray176},
ylabel={GPU Memory Usage (GiB)},
ymajorgrids,
ymin=0, ymax=9,
ytick style={color=black}
]
\addplot [thick, royalblue4102200, dashed]
table {%
0 1.69716548919678
1 3.22841548919678
2 4.75966548919678
3 3.61122798919678
4 2.84560298919678
5 3.22841548919678
6 2.65419673919678
7 3.03700923919678
8 3.22841548919678
9 3.22841548919678
10 3.22841548919678
11 2.94130611419678
12 3.13271236419678
13 3.22841548919678
14 3.22841548919678
15 3.22841548919678
16 3.08486080169678
17 3.13271236419678
18 3.13271236419678
19 3.13271236419678
20 3.13271236419678
21 3.09682369232178
22 3.09877681732178
23 3.08876705169678
24 3.08729076385498
25 3.1275634765625
26 3.1270866394043
27 3.1231803894043
28 3.0973014831543
29 3.4202995300293
30 3.3724479675293
31 3.3245964050293
32 3.2288932800293
33 3.0853385925293
34 4.0902214050293
35 3.8988151550293
36 3.7074089050293
37 3.3245964050293
38 2.9417839050293
39 4.9515495300293
40 4.5687370300293
41 4.1859245300293
42 3.4202995300293
43 2.6546745300293
44 8.2054557800293
45 6.6742057800293
46 6.6742057800293
47 6.2913932800293
48 3.2288932800293
49 1.69764280319214
};
\addlegendentry{$O(\sqrt{n})$ Algo.}
\addplot [thick, crimson2393560]
table {%
0 1.69838619232178
1 3.22963619232178
2 4.76088619232178
3 3.61244869232178
4 2.84682369232178
5 3.61244869232178
6 3.03822994232178
7 2.65541744232178
8 2.84682369232178
9 2.84682369232178
10 2.84682369232178
11 2.55971431732178
12 2.36830806732178
13 2.55971431732178
14 2.55971431732178
15 2.55971431732178
16 2.41615962982178
17 2.27260494232178
18 2.27260494232178
19 2.27260494232178
20 2.27260494232178
21 2.23671627044678
22 2.19081783294678
23 2.18080806732178
24 2.17933177947998
25 3.3209228515625
26 3.3204460144043
27 3.3158073425293
28 3.3377799987793
29 3.3258171081543
30 3.2779655456543
31 3.2301139831543
32 3.1344108581543
33 3.1822624206543
34 3.1344108581543
35 2.9430046081543
36 2.7515983581543
37 2.3687858581543
38 6.1969108581543
39 6.1012077331543
40 5.7183952331543
41 5.3355827331543
42 4.5699577331543
43 4.5699577331543
44 4.3785514831543
45 2.8473014831543
46 6.6754264831543
47 6.2926139831543
48 3.2301139831543
49 1.69886350631714
};
\addlegendentry{Our Algo.}
\end{axis}

\end{tikzpicture}

%% file: tikz/our_algo_vs_optimal_ACG.tex
% This file was created with tikzplotlib v0.10.1.
\begin{tikzpicture}

\definecolor{crimson2393560}{RGB}{239,35,60}
\definecolor{darkgray176}{RGB}{176,176,176}
\definecolor{lightgray204}{RGB}{204,204,204}
\definecolor{royalblue4102200}{RGB}{4,102,200}

\begin{axis}[
legend cell align={left},
legend style={fill opacity=0.8, draw opacity=1, text opacity=1, draw=lightgray204},
tick align=outside,
tick pos=left,
title={ },
x grid style={darkgray176},
xlabel={Training Phase Index},
xmajorgrids,
xmin=0, xmax=49,
xtick style={color=black},
xticklabel style={rotate=340.0},
ytick={1,2,3,4,5,6,7,8,9},
xtick={3,6,24,46,43,25},
xticklabel style={rotate=340.0, font=\tiny},
legend style={font=\scriptsize, at={(0,1)},anchor=north west, xshift=5pt,yshift=-5pt},
y grid style={darkgray176},
ylabel={GPU Memory Usage (GiB)},,
ymajorgrids,
ymin=0, ymax=9,
ytick style={color=black}
]
\addplot [thick, royalblue4102200, dashed]
table {%
0 1.69899654388428
1 3.23024654388428
2 4.76149654388428
3 3.61305904388428
4 2.84743404388428
5 3.61305904388428
6 3.03884029388428
7 2.65602779388428
8 3.03884029388428
9 3.03884029388428
10 3.03884029388428
11 2.75173091888428
12 2.56032466888428
13 2.65602779388428
14 2.65602779388428
15 2.65602779388428
16 2.51247310638428
17 2.36891841888428
18 2.36891841888428
19 2.36891841888428
20 2.36891841888428
21 2.33302974700928
22 2.28713130950928
23 2.27712154388428
24 2.27564525604248
25 5.23486328125
26 5.2343864440918
27 5.2312126159668
28 5.2531852722168
29 5.2404899597168
30 5.1926383972168
31 5.1447868347168
32 5.0490837097168
33 5.0969352722168
34 5.0490837097168
35 4.8576774597168
36 4.6662712097168
37 4.2834587097168
38 4.2834587097168
39 4.1877555847168
40 3.8049430847168
41 3.4221305847168
42 2.6565055847168
43 4.5805680847168
44 4.3891618347168
45 2.8579118347168
46 6.6960368347168
47 6.2932243347168
48 3.2307243347168
49 1.69947385787964
};
\addlegendentry{Optimal ACG}
\addplot [thick, crimson2393560]
table {%
0 1.69838619232178
1 3.22963619232178
2 4.76088619232178
3 3.61244869232178
4 2.84682369232178
5 3.61244869232178
6 3.03822994232178
7 2.65541744232178
8 2.84682369232178
9 2.84682369232178
10 2.84682369232178
11 2.55971431732178
12 2.36830806732178
13 2.55971431732178
14 2.55971431732178
15 2.55971431732178
16 2.41615962982178
17 2.27260494232178
18 2.27260494232178
19 2.27260494232178
20 2.27260494232178
21 2.23671627044678
22 2.19081783294678
23 2.18080806732178
24 2.17933177947998
25 3.3209228515625
26 3.3204460144043
27 3.3158073425293
28 3.3377799987793
29 3.3258171081543
30 3.2779655456543
31 3.2301139831543
32 3.1344108581543
33 3.1822624206543
34 3.1344108581543
35 2.9430046081543
36 2.7515983581543
37 2.3687858581543
38 6.1969108581543
39 6.1012077331543
40 5.7183952331543
41 5.3355827331543
42 4.5699577331543
43 4.5699577331543
44 4.3785514831543
45 2.8473014831543
46 6.6754264831543
47 6.2926139831543
48 3.2301139831543
49 1.69886350631714
};
\addlegendentry{Our Algo.}
\end{axis}

\end{tikzpicture}

%% file: tikz/alexnet_optimal.tex
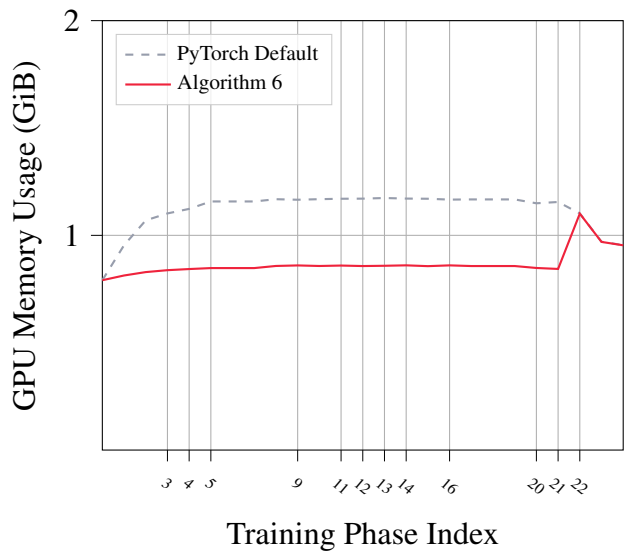
\begin{figure}[h!tb]
\centering

% This file was created with tikzplotlib v0.10.1.
\begin{tikzpicture}

\definecolor{crimson2393560}{RGB}{239,35,60}
\definecolor{darkgray151157172}{RGB}{151,157,172}
\definecolor{darkgray176}{RGB}{176,176,176}
\definecolor{lightgray204}{RGB}{204,204,204}

\begin{axis}[
legend cell align={left},
legend style={fill opacity=0.8, draw opacity=1, text opacity=1, draw=lightgray204},
tick align=outside,
tick pos=left,
title={ },
x grid style={darkgray176},
xlabel={Training Phase Index},
xmajorgrids,
xmin=0, xmax=24,
xtick style={color=black},
xticklabel style={rotate=340.0},
ytick={1,2},
xtick={3,4,5,9,11,12,13,14,16,20,21,22},
xticklabel style={rotate=340.0, font=\tiny},
legend style={font=\scriptsize, at={(0,1)},anchor=north west, xshift=5pt,yshift=-5pt},
y grid style={darkgray176},
ylabel={GPU Memory Usage (GiB)},,
ymajorgrids,
ymin=0, ymax=2,
ytick style={color=black}
]
\addplot [thick, darkgray151157172, dashed]
table {%
0 0.79103125
1 0.95390625
2 1.06978125
3 1.10209375
4 1.12321875
5 1.15784375
6 1.15784375
7 1.15784375
8 1.16834375
9 1.16584375
10 1.16846875
11 1.17046875
12 1.17095703125
13 1.1738837890625
14 1.1712705078125
15 1.1704580078125
16 1.1664580078125
17 1.1673330078125
18 1.1673330078125
19 1.1673330078125
20 1.1493330078125
21 1.1553955078125
22 1.1019580078125
23 0.9702392578125
24 0.9543955078125
};
\addlegendentry{PyTorch Default}
\addplot [thick, crimson2393560]
table {%
0 0.79103125
1 0.81303125
2 0.828875
3 0.83778125
4 0.8430625
5 0.8475625
6 0.8475625
7 0.8475625
8 0.8574375
9 0.85990625
10 0.857328125
11 0.859328125
12 0.85701953125
13 0.8583056640625
14 0.8603173828125
15 0.8565517578125
16 0.8601767578125
17 0.8570517578125
18 0.8570517578125
19 0.8570517578125
20 0.8480517578125
21 0.8435517578125
22 1.1011767578125
23 0.9694580078125
24 0.9536142578125
};
\addlegendentry{Algorithm~\ref{alg:dynamic-checkpoint-selection}}
\end{axis}
\end{tikzpicture}

    \caption{GPU Memory Usage Profiling on Plain AlexNet with The Optimal Set of Checkpoints, \{3, 4, 5, 9, 11, 12\}.} 
    \label{fig:GMU_AlexNet_Ckpt_Optimal}
\end{figure}

%% file: tikz/alexnet_optimal2.tex
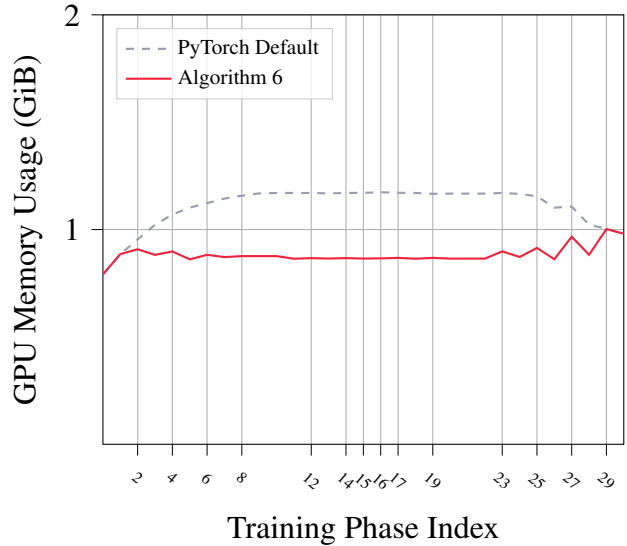
\begin{figure}[h!tb]
\centering

% This file was created with tikzplotlib v0.10.1.
\begin{tikzpicture}

\definecolor{crimson2393560}{RGB}{239,35,60}
\definecolor{darkgray151157172}{RGB}{151,157,172}
\definecolor{darkgray176}{RGB}{176,176,176}
\definecolor{lightgray204}{RGB}{204,204,204}

\begin{axis}[
legend cell align={left},
legend style={fill opacity=0.8, draw opacity=1, text opacity=1, draw=lightgray204},
tick align=outside,
tick pos=left,
title={ },
x grid style={darkgray176},
xlabel={Training Phase Index},
xmajorgrids,
xmin=0, xmax=30,
xtick style={color=black},
xticklabel style={rotate=340.0},
ytick={1,2},
xtick={2,4,6,8,12,14,15,16,17,19,23,25,27,29},
xticklabel style={rotate=340.0, font=\tiny},
legend style={font=\scriptsize, at={(0,1)},anchor=north west, xshift=5pt,yshift=-5pt},
y grid style={darkgray176},
ylabel={GPU Memory Usage (GiB)},,
ymajorgrids,
ymin=0, ymax=2,
ytick style={color=black}
]
\addplot [thick, darkgray151157172, dashed]
table {%
0 0.79103125
1 0.8855625
2 0.95390625
3 1.02225
4 1.06978125
5 1.10209375
6 1.12321875
7 1.14434375
8 1.15784375
9 1.16834375
10 1.17034375
11 1.17034375
12 1.17034375
13 1.16846875
14 1.17046875
15 1.17095703125
16 1.1738837890625
17 1.1712705078125
18 1.1704580078125
19 1.1664580078125
20 1.1673330078125
21 1.1673330078125
22 1.1673330078125
23 1.1704580078125
24 1.1659580078125
25 1.1553955078125
26 1.1019580078125
27 1.1069267578125
28 1.0227392578125
29 1.0033642578125
30 0.9805830078125
};
\addlegendentry{PyTorch Default}
\addplot [thick, crimson2393560]
table {%
0 0.79103125
1 0.88553125
2 0.9083125
3 0.882125
4 0.89796875
5 0.8613125
6 0.8824375
7 0.871875
8 0.876375
9 0.876375
10 0.876375
11 0.86425
12 0.866875
13 0.86503125
14 0.86703125
15 0.86486328125
16 0.8660087890625
17 0.8680205078125
18 0.8642392578125
19 0.8682392578125
20 0.8647392578125
21 0.8647392578125
22 0.8647392578125
23 0.8983173828125
24 0.8723642578125
25 0.9146142578125
26 0.8618017578125
27 0.9668017578125
28 0.8826142578125
29 1.0023330078125
30 0.9805517578125
};
\addlegendentry{Algorithm~\ref{alg:dynamic-checkpoint-selection}}
\end{axis}
\end{tikzpicture}

    \caption{GPU Memory Usage Profiling on Modified AlexNet with The Optimal Set of Checkpoints, \{2, 4, 6, 8, 12, 14, 15\}.} 
    \label{fig:GMU_AlexNet_Ckpt_Optimal2}
\end{figure}

%% file: 6_conclusion.tex
\section{Conclusion and Future Work}\label{sec:conclusion}

In this paper, we identify the memory pressure problem commonly seen in modern Deep Neural Network training among state-of-the-art platforms. Then we review the well-known insight proposed by Chen et al., i.e. {\em trading computation for memory}, and build a dynamic programming algorithm accordingly. From there, we refine the objective function for a more precise estimation of GPU memory cost to align our result with the prominent state-of-the-art deep-learning platform, PyTorch. With extensive experiments, we show that a more precise model specification can decrease peak GPU memory usage even more. While that means an increase in the input size, it is not a problem for our algorithm because it runs in linear time.

In the future, we would like to investigate the mechanism of GPU memory allocation in modern deep-learning platforms to reduce those {\em preserved memory} reported by Nvidia hardware. On the other hand, since we can divide an arbitrary computation graph into many linear sub-models, it will be interesting to generalize our algorithm for a general DNN model, including those with thousands of layers, which is apparently not analyzable by a naive $O(n^3)$ algorithm.

%% file: paper.bbl
\begin{thebibliography}{31}
\expandafter\ifx\csname natexlab\endcsname\relax\def\natexlab#1{#1}\fi
\providecommand{\url}[1]{\texttt{#1}}
\providecommand{\href}[2]{#2}
\providecommand{\path}[1]{#1}
\providecommand{\DOIprefix}{doi:}
\providecommand{\ArXivprefix}{arXiv:}
\providecommand{\URLprefix}{URL: }
\providecommand{\Pubmedprefix}{pmid:}
\providecommand{\doi}[1]{\href{http://dx.doi.org/#1}{\path{#1}}}
\providecommand{\Pubmed}[1]{\href{pmid:#1}{\path{#1}}}
\providecommand{\bibinfo}[2]{#2}
\ifx\xfnm\relax \def\xfnm[#1]{\unskip,\space#1}\fi
%Type = Inproceedings
\bibitem[{Beaumont et~al.(2021)Beaumont, Eyraud-Dubois and Shilova}]{Beaumont2021EfficientCO}
\bibinfo{author}{Beaumont, O.}, \bibinfo{author}{Eyraud-Dubois, L.}, \bibinfo{author}{Shilova, A.}, \bibinfo{year}{2021}.
\newblock \bibinfo{title}{Efficient combination of rematerialization and offloading for training dnns}, in: \bibinfo{booktitle}{Neural Information Processing Systems}.
%Type = Misc
\bibitem[{Chen et~al.(2016)Chen, Xu, Zhang and Guestrin}]{chen2016training}
\bibinfo{author}{Chen, T.}, \bibinfo{author}{Xu, B.}, \bibinfo{author}{Zhang, C.}, \bibinfo{author}{Guestrin, C.}, \bibinfo{year}{2016}.
\newblock \bibinfo{title}{Training deep nets with sublinear memory cost}.
\newblock \href{http://arxiv.org/abs/1604.06174}{{\tt arXiv:1604.06174}}.
%Type = Misc
\bibitem[{Feng and Huang(2021)}]{feng2021optimal}
\bibinfo{author}{Feng, J.}, \bibinfo{author}{Huang, D.}, \bibinfo{year}{2021}.
\newblock \bibinfo{title}{Optimal gradient checkpoint search for arbitrary computation graphs}.
\newblock \href{http://arxiv.org/abs/1808.00079}{{\tt arXiv:1808.00079}}.
%Type = Inproceedings
\bibitem[{Gomez et~al.(2017)Gomez, Ren, Urtasun and Grosse}]{Gomez2017TheRR}
\bibinfo{author}{Gomez, A.N.}, \bibinfo{author}{Ren, M.}, \bibinfo{author}{Urtasun, R.}, \bibinfo{author}{Grosse, R.B.}, \bibinfo{year}{2017}.
\newblock \bibinfo{title}{The reversible residual network: Backpropagation without storing activations}, in: \bibinfo{booktitle}{NIPS}.
%Type = Book
\bibitem[{Goodfellow et~al.(2016)Goodfellow, Bengio and Courville}]{Goodfellow-et-al-2016}
\bibinfo{author}{Goodfellow, I.}, \bibinfo{author}{Bengio, Y.}, \bibinfo{author}{Courville, A.}, \bibinfo{year}{2016}.
\newblock \bibinfo{title}{Deep Learning}.
\newblock \bibinfo{publisher}{MIT Press}.
\newblock \bibinfo{note}{\url{http://www.deeplearningbook.org}}.
%Type = Inproceedings
\bibitem[{Gruslys et~al.(2016)Gruslys, Munos, Danihelka, Lanctot and Graves}]{Gruslys2016MemoryEfficientBT}
\bibinfo{author}{Gruslys, A.}, \bibinfo{author}{Munos, R.}, \bibinfo{author}{Danihelka, I.}, \bibinfo{author}{Lanctot, M.}, \bibinfo{author}{Graves, A.}, \bibinfo{year}{2016}.
\newblock \bibinfo{title}{Memory-efficient backpropagation through time}, in: \bibinfo{booktitle}{NIPS}.
%Type = Inproceedings
\bibitem[{Gupta et~al.(2015)Gupta, Agrawal, Gopalakrishnan and Narayanan}]{Gupta2015DeepLW}
\bibinfo{author}{Gupta, S.}, \bibinfo{author}{Agrawal, A.}, \bibinfo{author}{Gopalakrishnan, K.}, \bibinfo{author}{Narayanan, P.}, \bibinfo{year}{2015}.
\newblock \bibinfo{title}{Deep learning with limited numerical precision}, in: \bibinfo{booktitle}{International Conference on Machine Learning}.
%Type = Article
\bibitem[{Han et~al.(2015a)Han, Mao and Dally}]{Han2015DeepCC}
\bibinfo{author}{Han, S.}, \bibinfo{author}{Mao, H.}, \bibinfo{author}{Dally, W.J.}, \bibinfo{year}{2015}a.
\newblock \bibinfo{title}{Deep compression: Compressing deep neural network with pruning, trained quantization and huffman coding}.
\newblock \bibinfo{journal}{arXiv: Computer Vision and Pattern Recognition} .
%Type = Inproceedings
\bibitem[{Han et~al.(2015b)Han, Pool, Tran and Dally}]{Han2015LearningBW}
\bibinfo{author}{Han, S.}, \bibinfo{author}{Pool, J.}, \bibinfo{author}{Tran, J.}, \bibinfo{author}{Dally, W.J.}, \bibinfo{year}{2015}b.
\newblock \bibinfo{title}{Learning both weights and connections for efficient neural network}, in: \bibinfo{booktitle}{NIPS}.
%Type = Article
\bibitem[{He et~al.(2015)He, Zhang, Ren and Sun}]{He2015DeepRL}
\bibinfo{author}{He, K.}, \bibinfo{author}{Zhang, X.}, \bibinfo{author}{Ren, S.}, \bibinfo{author}{Sun, J.}, \bibinfo{year}{2015}.
\newblock \bibinfo{title}{Deep residual learning for image recognition}.
\newblock \bibinfo{journal}{2016 IEEE Conference on Computer Vision and Pattern Recognition (CVPR)} , \bibinfo{pages}{770--778}.
%Type = Misc
\bibitem[{Herrmann et~al.(2019)Herrmann, Beaumont, Eyraud-Dubois, Hermann, Joly and Shilova}]{herrmann2019optimal}
\bibinfo{author}{Herrmann, J.}, \bibinfo{author}{Beaumont, O.}, \bibinfo{author}{Eyraud-Dubois, L.}, \bibinfo{author}{Hermann, J.}, \bibinfo{author}{Joly, A.}, \bibinfo{author}{Shilova, A.}, \bibinfo{year}{2019}.
\newblock \bibinfo{title}{Optimal checkpointing for heterogeneous chains: how to train deep neural networks with limited memory}.
\newblock \href{http://arxiv.org/abs/1911.13214}{{\tt arXiv:1911.13214}}.
%Type = Article
\bibitem[{Hu et~al.(2017)Hu, Shen, Albanie, Sun and Wu}]{Hu2017SqueezeandExcitationN}
\bibinfo{author}{Hu, J.}, \bibinfo{author}{Shen, L.}, \bibinfo{author}{Albanie, S.}, \bibinfo{author}{Sun, G.}, \bibinfo{author}{Wu, E.}, \bibinfo{year}{2017}.
\newblock \bibinfo{title}{Squeeze-and-excitation networks}.
\newblock \bibinfo{journal}{IEEE Transactions on Pattern Analysis and Machine Intelligence} \bibinfo{volume}{42}, \bibinfo{pages}{2011--2023}.
%Type = Article
\bibitem[{chin Huang et~al.(2020)chin Huang, Jin and Li}]{Huang2020SwapAdvisorPD}
\bibinfo{author}{chin Huang, C.}, \bibinfo{author}{Jin, G.}, \bibinfo{author}{Li, J.}, \bibinfo{year}{2020}.
\newblock \bibinfo{title}{Swapadvisor: Pushing deep learning beyond the gpu memory limit via smart swapping}.
\newblock \bibinfo{journal}{Proceedings of the Twenty-Fifth International Conference on Architectural Support for Programming Languages and Operating Systems} .
%Type = Inproceedings
\bibitem[{Judd et~al.(2016)Judd, Albericio, Hetherington, Aamodt, Jerger and Moshovos}]{10.1145/2925426.2926294}
\bibinfo{author}{Judd, P.}, \bibinfo{author}{Albericio, J.}, \bibinfo{author}{Hetherington, T.}, \bibinfo{author}{Aamodt, T.M.}, \bibinfo{author}{Jerger, N.E.}, \bibinfo{author}{Moshovos, A.}, \bibinfo{year}{2016}.
\newblock \bibinfo{title}{Proteus: Exploiting numerical precision variability in deep neural networks}, in: \bibinfo{booktitle}{Proceedings of the 2016 International Conference on Supercomputing}, \bibinfo{publisher}{Association for Computing Machinery}, \bibinfo{address}{New York, NY, USA}.
\newblock \URLprefix \url{https://doi.org/10.1145/2925426.2926294}, \DOIprefix\doi{10.1145/2925426.2926294}.
%Type = Inproceedings
\bibitem[{Kirisame et~al.(2021)Kirisame, Lyubomirsky, Haan, Brennan, He, Roesch, Chen and Tatlock}]{kirisame2021dynamic}
\bibinfo{author}{Kirisame, M.}, \bibinfo{author}{Lyubomirsky, S.}, \bibinfo{author}{Haan, A.}, \bibinfo{author}{Brennan, J.}, \bibinfo{author}{He, M.}, \bibinfo{author}{Roesch, J.}, \bibinfo{author}{Chen, T.}, \bibinfo{author}{Tatlock, Z.}, \bibinfo{year}{2021}.
\newblock \bibinfo{title}{Dynamic tensor rematerialization}, in: \bibinfo{booktitle}{International Conference on Learning Representations}.
\newblock \URLprefix \url{https://openreview.net/forum?id=Vfs_2RnOD0H}.
%Type = Article
\bibitem[{Krizhevsky et~al.(2017)Krizhevsky, Sutskever and Hinton}]{krizhevsky2017imagenet}
\bibinfo{author}{Krizhevsky, A.}, \bibinfo{author}{Sutskever, I.}, \bibinfo{author}{Hinton, G.E.}, \bibinfo{year}{2017}.
\newblock \bibinfo{title}{Imagenet classification with deep convolutional neural networks}.
\newblock \bibinfo{journal}{Communications of the ACM} \bibinfo{volume}{60}, \bibinfo{pages}{84--90}.
%Type = Article
\bibitem[{Le et~al.(2018)Le, Imai, Negishi and Kawachiya}]{Le2018TFLMSLM}
\bibinfo{author}{Le, T.D.}, \bibinfo{author}{Imai, H.}, \bibinfo{author}{Negishi, Y.}, \bibinfo{author}{Kawachiya, K.}, \bibinfo{year}{2018}.
\newblock \bibinfo{title}{Tflms: Large model support in tensorflow by graph rewriting}.
\newblock \bibinfo{journal}{ArXiv} \bibinfo{volume}{abs/1807.02037}.
%Type = Misc
\bibitem[{Paszke et~al.(2019)Paszke, Gross, Massa, Lerer, Bradbury, Chanan, Killeen, Lin, Gimelshein, Antiga, Desmaison, Köpf, Yang, DeVito, Raison, Tejani, Chilamkurthy, Steiner, Fang, Bai and Chintala}]{paszke2019pytorch}
\bibinfo{author}{Paszke, A.}, \bibinfo{author}{Gross, S.}, \bibinfo{author}{Massa, F.}, \bibinfo{author}{Lerer, A.}, \bibinfo{author}{Bradbury, J.}, \bibinfo{author}{Chanan, G.}, \bibinfo{author}{Killeen, T.}, \bibinfo{author}{Lin, Z.}, \bibinfo{author}{Gimelshein, N.}, \bibinfo{author}{Antiga, L.}, \bibinfo{author}{Desmaison, A.}, \bibinfo{author}{Köpf, A.}, \bibinfo{author}{Yang, E.}, \bibinfo{author}{DeVito, Z.}, \bibinfo{author}{Raison, M.}, \bibinfo{author}{Tejani, A.}, \bibinfo{author}{Chilamkurthy, S.}, \bibinfo{author}{Steiner, B.}, \bibinfo{author}{Fang, L.}, \bibinfo{author}{Bai, J.}, \bibinfo{author}{Chintala, S.}, \bibinfo{year}{2019}.
\newblock \bibinfo{title}{Pytorch: An imperative style, high-performance deep learning library}.
\newblock \href{http://arxiv.org/abs/1912.01703}{{\tt arXiv:1912.01703}}.
%Type = Article
\bibitem[{Pudipeddi et~al.(2020)Pudipeddi, Mesmakhosroshahi, Xi and Bharadwaj}]{Pudipeddi2020TrainingLN}
\bibinfo{author}{Pudipeddi, B.}, \bibinfo{author}{Mesmakhosroshahi, M.}, \bibinfo{author}{Xi, J.}, \bibinfo{author}{Bharadwaj, S.}, \bibinfo{year}{2020}.
\newblock \bibinfo{title}{Training large neural networks with constant memory using a new execution algorithm}.
\newblock \bibinfo{journal}{ArXiv} \bibinfo{volume}{abs/2002.05645}.
%Type = Article
\bibitem[{Rhu et~al.(2016)Rhu, Gimelshein, Clemons, Zulfiqar and Keckler}]{Rhu2016vDNNVD}
\bibinfo{author}{Rhu, M.}, \bibinfo{author}{Gimelshein, N.}, \bibinfo{author}{Clemons, J.}, \bibinfo{author}{Zulfiqar, A.}, \bibinfo{author}{Keckler, S.W.}, \bibinfo{year}{2016}.
\newblock \bibinfo{title}{vdnn: Virtualized deep neural networks for scalable, memory-efficient neural network design}.
\newblock \bibinfo{journal}{2016 49th Annual IEEE/ACM International Symposium on Microarchitecture (MICRO)} , \bibinfo{pages}{1--13}.
%Type = Article
\bibitem[{Russakovsky et~al.(2015)Russakovsky, Deng, Su, Krause, Satheesh, Ma, Huang, Karpathy, Khosla, Bernstein et~al.}]{russakovsky2015imagenet}
\bibinfo{author}{Russakovsky, O.}, \bibinfo{author}{Deng, J.}, \bibinfo{author}{Su, H.}, \bibinfo{author}{Krause, J.}, \bibinfo{author}{Satheesh, S.}, \bibinfo{author}{Ma, S.}, \bibinfo{author}{Huang, Z.}, \bibinfo{author}{Karpathy, A.}, \bibinfo{author}{Khosla, A.}, \bibinfo{author}{Bernstein, M.}, et~al., \bibinfo{year}{2015}.
\newblock \bibinfo{title}{Imagenet large scale visual recognition challenge}.
\newblock \bibinfo{journal}{International journal of computer vision} \bibinfo{volume}{115}, \bibinfo{pages}{211--252}.
%Type = Article
\bibitem[{Simonyan and Zisserman(2014)}]{Simonyan2014VeryDC}
\bibinfo{author}{Simonyan, K.}, \bibinfo{author}{Zisserman, A.}, \bibinfo{year}{2014}.
\newblock \bibinfo{title}{Very deep convolutional networks for large-scale image recognition}.
\newblock \bibinfo{journal}{CoRR} \bibinfo{volume}{abs/1409.1556}.
%Type = Misc
\bibitem[{Sohoni et~al.(2022)Sohoni, Aberger, Leszczynski, Zhang and Ré}]{sohoni2022lowmemory}
\bibinfo{author}{Sohoni, N.S.}, \bibinfo{author}{Aberger, C.R.}, \bibinfo{author}{Leszczynski, M.}, \bibinfo{author}{Zhang, J.}, \bibinfo{author}{Ré, C.}, \bibinfo{year}{2022}.
\newblock \bibinfo{title}{Low-memory neural network training: A technical report}.
\newblock \href{http://arxiv.org/abs/1904.10631}{{\tt arXiv:1904.10631}}.
%Type = Inproceedings
\bibitem[{Szegedy et~al.(2017)Szegedy, Ioffe, Vanhoucke and Alemi}]{10.5555/3298023.3298188}
\bibinfo{author}{Szegedy, C.}, \bibinfo{author}{Ioffe, S.}, \bibinfo{author}{Vanhoucke, V.}, \bibinfo{author}{Alemi, A.A.}, \bibinfo{year}{2017}.
\newblock \bibinfo{title}{Inception-v4, inception-resnet and the impact of residual connections on learning}, in: \bibinfo{booktitle}{Proceedings of the Thirty-First AAAI Conference on Artificial Intelligence}, \bibinfo{publisher}{AAAI Press}. p. \bibinfo{pages}{4278–4284}.
%Type = Article
\bibitem[{Tan and Le(2019)}]{Tan2019EfficientNetRM}
\bibinfo{author}{Tan, M.}, \bibinfo{author}{Le, Q.V.}, \bibinfo{year}{2019}.
\newblock \bibinfo{title}{Efficientnet: Rethinking model scaling for convolutional neural networks}.
\newblock \bibinfo{journal}{ArXiv} \bibinfo{volume}{abs/1905.11946}.
%Type = Article
\bibitem[{Wu et~al.(2016a)Wu, Schuster, Chen, Le, Norouzi, Macherey, Krikun, Cao, Gao, Macherey, Klingner, Shah, Johnson, Liu, Kaiser, Gouws, Kato, Kudo, Kazawa, Stevens, Kurian, Patil, Wang, Young, Smith, Riesa, Rudnick, Vinyals, Corrado, Hughes and Dean}]{Wu2016GooglesNM}
\bibinfo{author}{Wu, Y.}, \bibinfo{author}{Schuster, M.}, \bibinfo{author}{Chen, Z.}, \bibinfo{author}{Le, Q.V.}, \bibinfo{author}{Norouzi, M.}, \bibinfo{author}{Macherey, W.}, \bibinfo{author}{Krikun, M.}, \bibinfo{author}{Cao, Y.}, \bibinfo{author}{Gao, Q.}, \bibinfo{author}{Macherey, K.}, \bibinfo{author}{Klingner, J.}, \bibinfo{author}{Shah, A.}, \bibinfo{author}{Johnson, M.}, \bibinfo{author}{Liu, X.}, \bibinfo{author}{Kaiser, L.}, \bibinfo{author}{Gouws, S.}, \bibinfo{author}{Kato, Y.}, \bibinfo{author}{Kudo, T.}, \bibinfo{author}{Kazawa, H.}, \bibinfo{author}{Stevens, K.}, \bibinfo{author}{Kurian, G.}, \bibinfo{author}{Patil, N.}, \bibinfo{author}{Wang, W.}, \bibinfo{author}{Young, C.}, \bibinfo{author}{Smith, J.R.}, \bibinfo{author}{Riesa, J.}, \bibinfo{author}{Rudnick, A.}, \bibinfo{author}{Vinyals, O.}, \bibinfo{author}{Corrado, G.S.}, \bibinfo{author}{Hughes, M.}, \bibinfo{author}{Dean, J.}, \bibinfo{year}{2016}a.
\newblock \bibinfo{title}{Google's neural machine translation system: Bridging the gap between human and machine translation}.
\newblock \bibinfo{journal}{ArXiv} \bibinfo{volume}{abs/1609.08144}.
%Type = Article
\bibitem[{Wu et~al.(2016b)Wu, Shen and van~den Hengel}]{Wu2016HighperformanceSS}
\bibinfo{author}{Wu, Z.}, \bibinfo{author}{Shen, C.}, \bibinfo{author}{van~den Hengel, A.}, \bibinfo{year}{2016}b.
\newblock \bibinfo{title}{High-performance semantic segmentation using very deep fully convolutional networks}.
\newblock \bibinfo{journal}{ArXiv} \bibinfo{volume}{abs/1604.04339}.
%Type = Article
\bibitem[{Yu et~al.(2022)Yu, Wang, Vasudevan, Yeung, Seyedhosseini and Wu}]{Yu2022CoCaCC}
\bibinfo{author}{Yu, J.}, \bibinfo{author}{Wang, Z.}, \bibinfo{author}{Vasudevan, V.}, \bibinfo{author}{Yeung, L.}, \bibinfo{author}{Seyedhosseini, M.}, \bibinfo{author}{Wu, Y.}, \bibinfo{year}{2022}.
\newblock \bibinfo{title}{Coca: Contrastive captioners are image-text foundation models}.
\newblock \bibinfo{journal}{Trans. Mach. Learn. Res.} \bibinfo{volume}{2022}.
%Type = Article
\bibitem[{{Zagoruyko} and {Komodakis}(2016)}]{2016arXiv160507146Z}
\bibinfo{author}{{Zagoruyko}, S.}, \bibinfo{author}{{Komodakis}, N.}, \bibinfo{year}{2016}.
\newblock \bibinfo{title}{{Wide Residual Networks}}.
\newblock \bibinfo{journal}{arXiv e-prints} , \bibinfo{pages}{arXiv:1605.07146}\DOIprefix\doi{10.48550/arXiv.1605.07146}, \href{http://arxiv.org/abs/1605.07146}{{\tt arXiv:1605.07146}}.
%Type = Inproceedings
\bibitem[{Zhu et~al.(2017)Zhu, Park, Isola and Efros}]{8237506}
\bibinfo{author}{Zhu, J.Y.}, \bibinfo{author}{Park, T.}, \bibinfo{author}{Isola, P.}, \bibinfo{author}{Efros, A.A.}, \bibinfo{year}{2017}.
\newblock \bibinfo{title}{Unpaired image-to-image translation using cycle-consistent adversarial networks}, in: \bibinfo{booktitle}{2017 IEEE International Conference on Computer Vision (ICCV)}, pp. \bibinfo{pages}{2242--2251}.
\newblock \DOIprefix\doi{10.1109/ICCV.2017.244}.
%Type = Article
\bibitem[{Zoph et~al.(2017)Zoph, Vasudevan, Shlens and Le}]{Zoph2017LearningTA}
\bibinfo{author}{Zoph, B.}, \bibinfo{author}{Vasudevan, V.}, \bibinfo{author}{Shlens, J.}, \bibinfo{author}{Le, Q.V.}, \bibinfo{year}{2017}.
\newblock \bibinfo{title}{Learning transferable architectures for scalable image recognition}.
\newblock \bibinfo{journal}{2018 IEEE/CVF Conference on Computer Vision and Pattern Recognition} , \bibinfo{pages}{8697--8710}.

\end{thebibliography}
